\documentclass[10pt,twocolumn,letterpaper]{article}

\usepackage{cvpr}
\usepackage{times}
\usepackage{epsfig}
\usepackage{graphicx}
\usepackage{amsmath}
\usepackage{amssymb}
\usepackage{authblk}

\usepackage{algpseudocode}
\usepackage{algorithm}
\usepackage{multirow}
\usepackage{booktabs}
\usepackage{subfigure}
\usepackage{colortbl}
\usepackage{tabularx}
\usepackage{appendix}
\usepackage[export]{adjustbox}
\usepackage[font=small,labelfont=bf]{caption}
\definecolor{LightCyan}{rgb}{0.88,1,1}
\definecolor{Gray}{gray}{0.9}

\newcommand*{\skipnumber}[2][1]{%
   {\renewcommand*{\alglinenumber}[1]{}\State #2}%
   \addtocounter{ALG@line}{-#1}}
 \newcommand{\nonl}{\renewcommand{\nl}{\let\nl\oldnl}}

\usepackage[breaklinks=true,bookmarks=false]{hyperref}

\cvprfinalcopy 

\def\cvprPaperID{****} 

\begin{document}

\title{CatNet: Class Incremental 3D ConvNets for Lifelong Egocentric Gesture Recognition}


\author{Zhengwei Wang$^1$\thanks{This work is finacially supported by Science Foundation Ireland (SFI) under the Grant Number 15/RP/2776.}, \hspace{5pt}Qi She$^2$\thanks{Corresponding author}, \hspace{5pt}Tejo Chalasani$^1$, and\hspace{5pt}Aljosa Smolic$^1$\\
\textsuperscript{1}V-SENSE, Trinity College Dublin \qquad \hspace{30pt} \textsuperscript{2}Intel Labs China\\
{\tt\small\{zhengwei.wang,CHALASAT,SMOLICA\}@tcd.ie \qquad qi.she@intel.com}
}

\maketitle

\begin{abstract}
Egocentric gestures are the most natural form of communication for humans to interact with wearable devices such as VR/AR helmets and glasses. A major issue in such scenarios for real-world applications is that may easily become necessary to add new gestures to the system e.g., a proper VR system should allow users to customize gestures incrementally. Traditional deep learning methods require storing all previous class samples in the system and training the model again from scratch by incorporating previous samples and new samples, which costs humongous memory and significantly increases computation over time. In this work, we demonstrate a lifelong 3D convolutional framework -- c(\textbf{C})la(\textbf{a})ss increment(\textbf{t})al net(\textbf{Net})works (CatNet), which considers temporal information in videos and enables lifelong learning for egocentric gesture video recognition by learning the feature representation of an exemplar set selected from previous class samples. Importantly, we propose a two-stream CatNet, which deploys RGB and depth modalities to train two separate networks. We evaluate CatNets on a publicly available dataset -- EgoGesture dataset, and show that CatNets can learn many classes incrementally over a long period of time. Results also demonstrate that the two-stream architecture achieves the best performance on both joint training and class incremental training compared to 3 other one-stream architectures. The codes and pre-trained models used in this work are provided at \url{https://github.com/villawang/CatNet}.
\end{abstract}

\vspace{-20pt}
\section{Introduction}
\vspace{-5pt}
With development and popularity of VR/AR devices recently, 
there is an increasing demand to work with these devices intuitively. Gestures are the most natural form for humans to interact with such type of devices, in which hand gestures can be conveniently captured by cameras integrated in the devices in first person view. This motivates accurate recognition of meaningful gestures from such egocentric gesture videos. 

Video recognition systems for such VR/AR applications in the real world should ideally be designed in a way to support incremental update and customization of gestures. Different communicative gestures should be customized for different VR games~\cite{yang2019gesture}. Traditional machine learning/deep learning approaches require training data of all classes accessed at the same time, which is hardly achievable in such real-world situations. For instance, when a new gesture should be added to a system, the model needs to be retrained by incorporating the gesture video samples of previous and new classes, which requires significant memory for storing all previous class videos and increasing computational cost over time. A system with capability of lifelong learning would therefore be very beneficial for such scenarios, in which incremental learning makes use of memory efficiently, enables fast learning for new class samples and does not forget the previous class samples. In this work, we demonstrate a c(\textbf{C})la(\textbf{a})ss increment(\textbf{t})al net(\textbf{Net})works (CatNet) for an open-set problem rather than a close-set problem, which learns new classes i.e., the class variants larger than instance variants.

\begin{figure}[!ht]
    \centering
    \includegraphics[width=.105\textwidth]{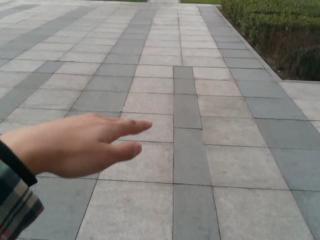}
    \includegraphics[width=.105\textwidth]{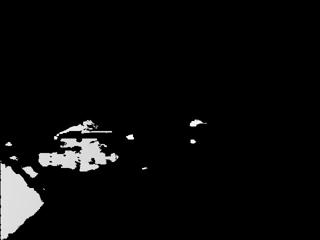}
    \hspace{10pt}
    \includegraphics[width=.105\textwidth]{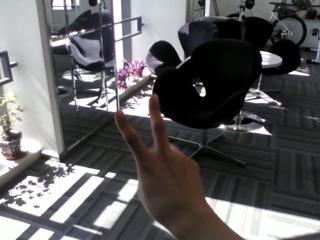}
    \includegraphics[width=.105\textwidth]{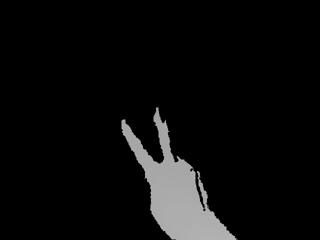}
    \caption{Visualization of gestures in different scenes. Left: The participant in a walking state with a dynamic background. Right: The participant in a stationary state facing a window with drastically changing sunlight.}
    \label{fig:frame_viz}
\end{figure}

Hand-crafted features are commonly adopted in traditional video gesture recognition~\cite{liu2013learning,ohn2014hand,wan2016chalearn}. With more large-scale datasets being released and development of deep neural networks (DNNs), DNNs are playing a more and more important role in this field~\cite{kopuklu2019real,cao2017egocentric}. Different from image recognition, temporal information along each frame needs to be considered for video understanding. The 3D convolutional network (ConvNet) becomes a popular architecture for learning spatiotemporal features from video clips. Benefiting from large-scale video datasets being released~\cite{kay2017kinetics,caba2015activitynet,abu2016youtube}, deep 3D ConvNets have achieved striking results in video action recognition tasks~\cite{hara2018can,carreira2017quo,diba2017temporal}. Compared to current popular video action datasets e.g., UCF-101~\cite{soomro2012ucf101}, Kinetics~\cite{kay2017kinetics}, egocentric gesture video is in first person view, in which two modalities RGB and depth can be captured at the same time. This indicates more information can be used to train the models for the egocentric gesture video recognition. Two-stream 3D ConvNets~\cite{simonyan2014two} is proposed for video action recognition by using optical flow~\cite{dosovitskiy2015flownet} in addition to RGB frames but optical flow is difficult to compute and to use for large-scale datasets~\cite{diba2017temporal}. We evaluate our models on a recently released large-scale egocentric gesture video dataset named EgoGesture~\cite{zhang2018egogesture}, in which RGB and depth modalities are provided. Benefiting from RGB-D video, we propose a two-stream architecture that deploys RGB and depth as two streams for egocentric gesture video recognition in this work, which deals the inconsistent quality of RGB and depth frames (see Figure~\ref{fig:frame_viz}) across different scenes (6 different scenes are included in the dataset) during the recording. Figure~\ref{fig:frame_viz} shows two gesture examples in two scenes respectively i.e., in a walking state with a dynamic background on the left and in a stationary state facing a window with drastically changing sunlight on the right. It can be noticed that the quality of the RGB input and the depth input are not consistent i.e., walking and outdoor capture can result in poor depth data, while illumination changes from changing sunlight can affect distribution of RGB pixels. Fusing features produced by a two-stream architecture can mitigate this issue, which results in a better overall performance. Previous work has shown that the frame-based approaches (e.g., VGG-16) are ineffective for the EgoGesture video recognition~\cite{zhang2018egogesture} because such methods do not take account into temporal information. Video-based approaches are required for accurate recognition in this scenario.

Significant advances have been made recently in computer vision and deep learning tasks including object recognition, detection, segmentation, etc. However, most of the models can only be trained in a batch setting, in which training data of all object classes is required for training the model in a roll. Lifelong learning~\cite{parisi2019continual} is a strategy to enable training the model continuously. To overcome the issue addressed earlier in the context of egocentric video recognition, in which the system should be able to learn new gestures incrementally, we introduce a lifelong learning framework -- c(\textbf{C})la(\textbf{a})ss increment(\textbf{t})al net(\textbf{Net})works (CatNet), which is specifically designed for lifelong egocentric gesture video recognition based on 3D convolutional networks (ConvNets). Importantly, we propose a two-stream CatNet using RGB and depth input as separate streams, which achieves the best performance in the class incremental learning task.

To summarize, our contribution are three-fold: 
\begin{itemize}
    \item To the best of our knowledge, we are the first to address the class incremental issues in the area of egocentric gesture video recognition and introduce the lifelong learning approaches to this area.
    
    \item We propose a two-stream CatNet for egocentric gesture video recognition, which treats RGB and depth as two separate streams and this type of CatNet is shown to perform best in the class incremental task.
    
    \item Our results show that CatNets can learn many classes incrementally over a long period of time i.e., the highest mean accuracy of presented CatNet has achieved 0.884.
\end{itemize}

\section{Related Work}
We introduce some recent literature with respect to video action recognition, EgoGesture video recognition and lifelong learning in this section.
\subsection{Video Action Recognition}\label{sec:video-action-recognition}
The success of convolutional networks (ConvNets) in object detection~\cite{redmon2016you}, object recognition~\cite{krizhevsky2012imagenet}, panoptic segmentation~\cite{long2015fully} tasks etc. has attracted growing interest for deploying them to other areas of computer vision. Video understanding has became a very popular research area recently, which is driven by several released large-scale datasets such as Kinetics~\cite{kay2017kinetics}, YouTube-8M~\cite{abu2016youtube}, ActivityNet~\cite{caba2015activitynet} and Sports-1M~\cite{karpathy2014large}. Unlike image tasks, video tasks require not only spatial information for each frame but also temporal information for neighboring frames, which poses a challenge for traditional methods performing on image tasks. Video understanding for untrimmed video datasets e.g., ActivityNet is still very challenging today because it requires to consider the possibility of accomplishing additional tasks such as untrimmed action classification and detection. Work discussed in this paper only considers the trimmed video scenario. 

Many methods have been proposed for video action recognition by introducing temporal information to the model. 3D convolution has been firstly introduced in~\cite{ji20123d}, which enables 3D convolutional networks (3D ConvNets) to extract features from both spatial and temporal dimensions. With the success in learning spatiotemporal information from consecutive frames by using 3D convolutional modules, several 3D types of architectures have been proposed in this field e.g., I3D~\cite{carreira2017quo}, P3D~\cite{qiu2017learning}, T3D~\cite{diba2017temporal} and R3D~\cite{hara2018can}. The work in \cite{hara2018can} addresses that it is important to use a pretrained model that is trained on a large-scale video dataset for a specific video task, which is able to avoid issues such as overfitting, difficult to converge and long time for training. The authors also demonstrate the efficacy of using R3D (use ResNet block as backbone for 3D convolution) for video action recognition, providing good performance and flexible architectures. 

By using more than one modality for video action recognition, multimodal representation has achieved remarkable results~\cite{wang2016exploring,simonyan2014two,carreira2017quo,feichtenhofer2016convolutional,wang2016temporal}. A typical architecture is the two-stream ConvNet~\cite{simonyan2014two,carreira2017quo}, which uses RGB frames and optical flow~\cite{dosovitskiy2015flownet} for training two separate networks. However, the computation of optical flow is very expensive, which limits its deployment in practice~\cite{diba2017temporal}. There are lots of depth cameras available on the market with acceptable price e.g., RealSense Camera SR300, which makes RGB frames and depth maps conveniently accessible for the egocentric-like datasets e.g., EgoGesture. In this work, we apply a two-stream 3D ConvNet by using RGB frames and depth frames, where the R3D is used as the backbone for our 3D ConvNet. 

\subsection{Egocentric Gesture Video Recognition} 
Datasets Like EgoGesture \cite{cao2017egocentric, zhang2018egogesture}, GreenScreen \cite{tejo2018} pave the wave for end-to-end learnable DNN architectures to address large-scale egocentric gesture recognition problems. Cao \etal \cite{cao2017egocentric} propose a neural network architecture by using a 3D ConvNet in tandem with spatiotemporal transformer modules and a LSTM for recognizing egocentric gestures from trimmed egocentric videos. In their network design, conceptually 3D ConvNets calculate the motion features and STTMs compensate for the ego motion. Shi \etal \cite{she2019openloris} improve on this approach by replacing spatiotemporal transformer modules with spatiotemporal deformable modules to overcome the issue of non-availability of local geometric transformations. 

Chalasani and Smolic \cite{tejo2019} propose a different network architecture that extracts embeddings specific to ego hands which are calculated as output from their encoder and decoder based architecture, which simultaneously computes hand segmentation. The embeddings thus generated for each trimmed video are then used in LSTMs to discern the gesture present in the video.

In a different approach, Abavisani \etal \cite{abavisani2019improving} propose a training strategy to use knowledge from multi-modal data to get better performance on unimodal 3D ConvNets. Unlike Cao \etal \cite{cao2017egocentric}, they train a separate network for each available modality and use a new spatiotemporal semantic alignment loss function, which they propose to share the knowledge among all the trained networks.

The scope for application of recognizing gestures from trimmed videos is limited. To address this issue, K{\"o}p{\"u}kl{\"u} \etal \cite{kopuklu2019real} introduce a network architecture that could enable offline working CNN based networks to work online using a sliding window approach. 

However, the idea of lifelong learning for egogesture recognition has not been explored in any of the mentioned papers. Given a new gesture, the entire network has to be trained with all the gestures starting the training process from the beginning, which becomes cumbersome as the number of gestures increases incrementally.

\subsection{Lifelong Learning}
Current state-of-the-art DNNs have achieved impressive performance on a variety of individual tasks. However, it still remains a substantial challenge for deep learning, which is learning multiple tasks continuously. When training DNNs on a new task, a standard DNN forgets most of the information related to previously learned tasks. This phenomenon is known as ``catastrophic forgetting''~\cite{mccloskey1989catastrophic}.

There are three scenarios in the area of lifelong learning~\cite{van2019three}: (1) Task incremental learning, where the task ID is provided during testing; (2) Domain incremental learning, where the task ID is not provided during testing and the model does not have to infer the task ID; and (3) Class incremental learning, the task ID is not provided during testing and the model has to infer the task ID. The first scenario is the easiest one and the model is always informed about which task is going to be performed. In this case, the model can be trained with task-specific components. A typical network for such a scenario can have a ``multi-headed'' output layer for each task and the rest of the model can be shared across tasks~\cite{van2019three}. A typical example for the second scenario is that the environment is changing e.g., image background changes but the objects remain the same for an object recognition task. The model has to solve the task but does not infer how the environment changes~\cite{feng2019challenges}. The last scenario is the most challenging one which requires the model to infer each task. For example, the model has to learn new classes of objects incrementally in an object recognition task. In this work, we focus on the most challenging scenario -- class incremental learning, where we address the importance for learning gesture classes incrementally regarding the egocentric gesture video recognition. 

Catastrophic forgetting appears when the new instance is significantly different from previous observed examples. Current strategies such as replay of old samples~\cite{gepperth2016bio,rebuffi2017icarl} and  regularization~\cite{benna2016computational,fusi2005cascade} can be deployed to mitigate this problem. FearNet was proposed in~\cite{kemker2017fearnet}, where a generative neural network~\cite{parisi2019continual,wang2019generative,wang2019neuroscore,wang2020use} is used to create pseudo-samples that are intermixed with recently observed examples stored in its hippocampal network. PathNet~\cite{fernando2017pathnet} was proposed as an ensemble method, where a generic algorithm is used to find the optimal path through a neural network of fixed size for replication and mutation. Ideally, the lifelong learning should be triggered by the availability of short videos of single objects and performed online on the hardware with fine-grained updates, while the mainstream of methods we study are limited with much lower temporal precision as our previous sequential learning models~\cite{she2018reduced,she2019neural}. In~\cite{rebuffi2017icarl}, iCaRL was proposed to cache the most representative samples from previous classes by using representation learning, which demonstrates good performance on class incremental learning. It is also easy to be extended to any type of network architectures. Benefiting from these advantages, we incorporate iCaRL into our CatNet to realize a lifelong learning system for egocentric gesture video recognition.

\section{Methodology}
In this section, we first elaborate on the type of 3D ConvNets investigated in this work, which is known as R3D. Then we present a two-stream 3D ConvNet for egocentric gesture video recognition (EgoGesture dataset is used in this work). Finally we introduce a CatNet, which incorporates the class incremental learning strategy with 3D ConvNets. Two evaluation metrics are presented at the end of this section.

\subsection{Architectures}
Two types of R3D architectures are investigated in this work, which use ResNet and ResNeXt respectively as the block unit. The difference between ResNet and ResNeXt is refered in~\cite{hara2018can}. Three models are studied, which are ResNet-50 using 16 frames as an input (ResNet-50-16f), ResNeXt-101 using 16 frames as an input (ResNeXt-101-16f), and ResNeXt-101 using 32 frames as an input (ResNeXt-101-32f)~\cite{hara2018can,kopuklu2019real}.  

\begin{figure*}
    \begin{minipage}{0.65\linewidth}
    \subfigure[Comparison between 2D and 3D convolutional operations. Left: 2D convolution on an image, in which the output is an image. Middle: 2D convolution on a video clip, in which the output is an image. Right: 3D convolution on a video clip, in which the output is also a video clip. Figure from~\cite{tran2015learning}.]{\includegraphics[width=0.55\textwidth]{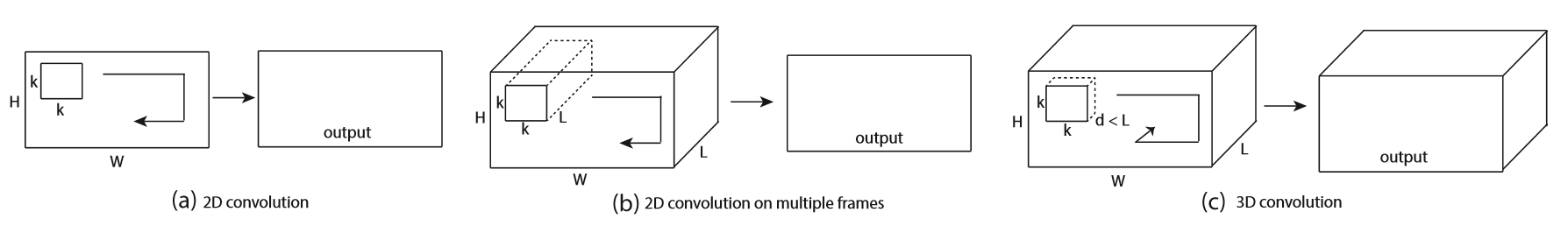}\label{fig:3D-convolutional-operation}}
    \hspace{5pt}
    \subfigure[Previous two-stream video architectures in the literature. K stands for the total number of frames in a video while N stands for a subset of neighboring frames of the video. Figure from~\cite{carreira2017quo}.]{\includegraphics[width=0.55\textwidth]{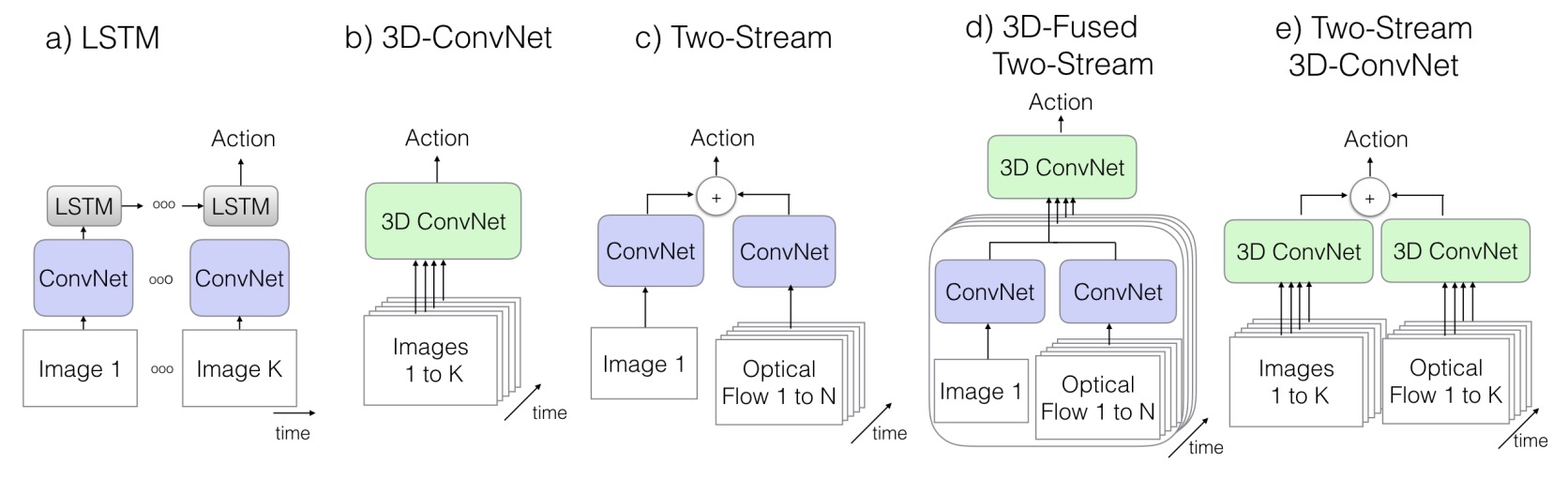}\label{fig:previous-3D-net}}
    \end{minipage}
    \hspace{-.8cm}
    \begin{minipage}{0.5\linewidth}
    \centering
    \subfigure[Two-stream 3D ConvNet used in this work. ]{\includegraphics[width=0.5\textwidth]{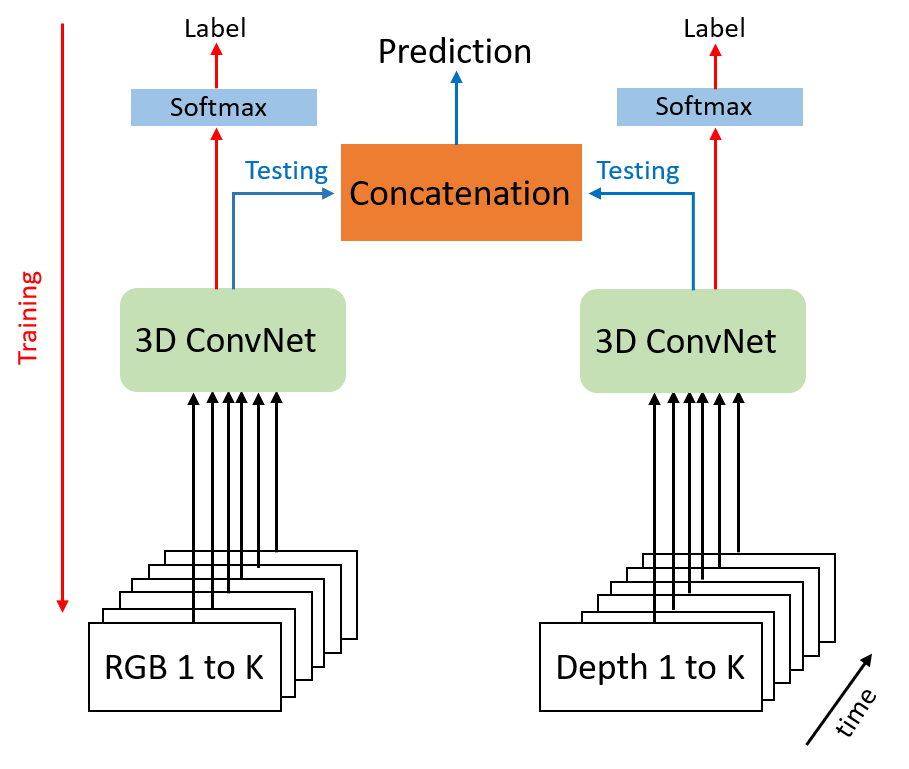}\label{fig:two-stream-model}}
    \end{minipage}
    \caption{Illustration of 3D ConvNets and the two-stream 3D ConvNet used in this work.}
    \label{fig:Illustration-3D-convolutional-networks}
\end{figure*}

As mentioned earlier in the paper, temporal information is important for video understanding. 3D convolution has became a popular operation to preserve the temporal properties of a video. Figure~\ref{fig:3D-convolutional-operation} illustrates the difference between the 3D convolutional operation and the 2D convolutional operation. 2D ConvNets lose track of temporal information of the input after every convolutional operation while 3D ConvNets are able to output a video clip by feeding a video clip, which preserves the temporal information. Figure~\ref{fig:previous-3D-net} illustrates previous popular architectures for video action recognition. Previous two-stream architectures learn the temporal information by using 3D convolution and optical flow~\cite{simonyan2014two,feichtenhofer2016convolutional,carreira2017quo,dosovitskiy2015flownet}. Optical flow represents the motion over time can be calculated from every two neighboring frames. Traditional two-stream architectures all use RGB frames and optical flow as two streams. However, the computation of optical flow is very complex i.e., computing over each individual frame, which is difficult in real-world applications~\cite{diba2017temporal}. We propose the use of depth frames as another stream with RGB frames for the EgoGesture dataset. It should be noticed that, differing from optical flow, the objective of the depth stream is not to extract temporal information. It aims to provide the depth level, in which different backgrounds e.g., brightness, indoor and outdoor in EgoGesture may drive different effects on RGB frames. The temporal information can be preserved by using 3D convolution. Figure~\ref{fig:two-stream-model} shows the two-stream architecture deployed in this work. Two 3D ConvNets are trained independently by using RGB and depth videos (see the training flow in Figure~\ref{fig:two-stream-model}) and the second last layer features of two networks are concatenated with each other (see the testing flow in Figure~\ref{fig:two-stream-model}), which is used for clustering during testing (we will explain this in the next section). 


\subsection{CatNet}
\begin{figure}[!ht]
    \centering
    \includegraphics[width=0.45\textwidth]{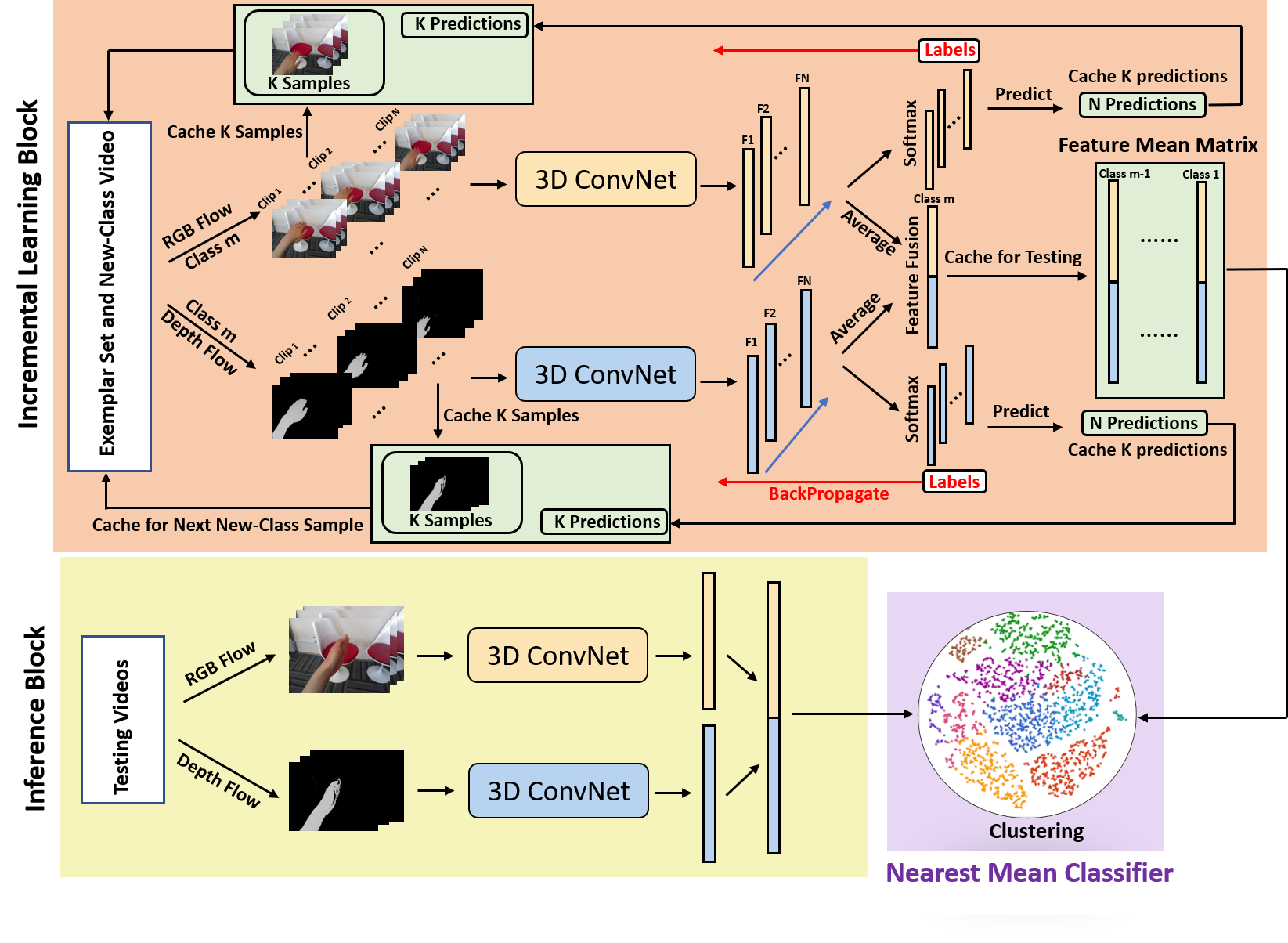}
    \caption{Schematic of a two-stream CatNet for EgoGesture video recognition.}
    \label{fig:testing_schematic}
\end{figure}

We incorporate iCaRL~\cite{rebuffi2017icarl} with 3D ConvNets for class incremental EgoGesture video recognition in this work and we call this framework CatNet. The whole training process for a CatNet is summarized in Algorithm~\ref{al:icarl} and Algorithm~\ref{al:icarl-cache-sample} and is visualized in Figure~\ref{fig:testing_schematic}. The core part of CatNet is to cache some previous class samples that are the most representative of the old class i.e., see the green block in Figure~\ref{fig:testing_schematic}. The memory caches the selected video samples and their corresponding predictions for previous class samples, which is achieved by learning the \textbf{feature representation} (Algorithm~\ref{al:icarl-cache-sample}). The feature representation is computed by the mean value of features (i.e., the second last layer output of the 3D ConvNet) corresponding to one class (see the feature mean matrix in Figure~\ref{fig:testing_schematic}). We then cache the first $k$ samples in which features of those samples are the closest to the representation (feature mean). The cached samples play two roles during the class incremental learning. First, the cached samples are used to compute the representation for each class, which is used for inference (see the nearest mean classifier in Figure~\ref{fig:testing_schematic}). Second, the prediction is used to compute the distillation loss during the training in Algorithm~\ref{al:icarl}. The inference procedure is summarized in Algorithm~\ref{al:icarl-inference} and the yellow block Figure~\ref{fig:testing_schematic}. The feature is extracted from a testing video, which is to be compared with the cached feature mean matrix. The class minimizing the $L_2$ distance is assigned as the predicted class. \textit{All features mentioned in this work are $L_2$-normalized}.
\begin{algorithm}[!ht]
    \caption{Training a CatNet for EgoGesture}
    \label{al:icarl}
    \begin{algorithmic}[1]
        \item[\textbf{Input:}]
        \begin{itemize}
            \Statex \item $m$ \Comment \textit{Number of added new classes}
            \Statex \item $\mathbf{X} \in \mathbb{R}^{N \times C \times L \times H \times W}$, $\mathbf{Y} \in \mathbb{R}^{N \times m}$ \Comment \textit{New- class video clips and labels, $N$ number of clips, $C$ number of channels, $L$ clip length, $H$ frame height, $W$ frame width}
            \Statex \item $K$ cached samples for each previous class \Comment \textit{$\mathbf{X}_{cached} \in \mathbb{R}^{P \times C \times L \times H \times W}$ where $P = K \times n$, $n$ is number of learned classes}
        \end{itemize}
        \item[\textbf{Require:}]
        \begin{itemize}
            \Statex \item Current model $\mathcal{M}$ and weights $\mathbf{\Theta}$ \Comment \textit{We denote the first layer weight until the last layer weight as $\Theta_1, \Theta_2, \ldots, \Theta_t$}
        \end{itemize}
        \item[\textbf{Training starts:}]
            \Statex $q = \mathcal{M}(\mathbf{X}_{cached}, \Theta_{1\sim t})$ \Comment \textit{Softmax prediction for previous samples}
            \Statex Optimizing (e.g., BackProp) with loss function below: 
            \Statex $\mathcal{L} = $ 
            \Statex $-\sum_{x_i\in X, y_i\in Y}\sum_{j=1}^m y_{i,j}\log(\mathcal{M}(x_i, \Theta_{1\sim t}))- $
            \Statex $\sum_{x_i\in X_{cached}, q_i\in q}\sum_{j=1}^P q_{i,j}\log(\mathcal{M}(x_i, \Theta_{1\sim t}))$ \Comment \textit{This contains the new-class cross entropy loss and the old-class distillation loss.}
        \item[\textbf{Training finishes}]
    \end{algorithmic}
\end{algorithm}

\begin{algorithm}[!ht]
    \caption{Learning the Feature Representation}
    \label{al:icarl-cache-sample}
    \begin{algorithmic}[1]
     \item[\textbf{Input:}]
        \begin{itemize}
            \Statex \item $\mathbf{X} \in \mathbb{R}^{N \times C \times L \times H \times W}$ \Comment \textit{New-class video clips}
        \end{itemize}
        \item[\textbf{Repeat for $m$ classes:}]
        \Statex $\mathbf{X}_i \in \mathbf{X}$ \Comment \textit{Samples of one new class}
        \Statex $\mathcal{F} = \mathcal{M}(\mathbf{X}_i, \Theta_{1\sim t-1})$ \Comment \textit{Extract the second last layer feature for one new-class samples}
        \Statex $\mu \gets \frac{1}{|\mathcal{F}|}\sum_{\mathcal{F}_i\in \mathcal{F}} \mathcal{F}_i$
        \skipnumber[3]{\For{$k = 1:K$} 
             \State $p_k \gets \mathrm{argmin}_{x \in X_i} \| \mu-\frac{1}{k}(\mathcal{M}(x, \Theta_{1\sim t-1})+\sum_{j=1}^{k-1}\mathcal{M}(p_j, \Theta_{1\sim t-1}) \|$
        \EndFor}
        \Statex $\mathbf{X}_{cached} \gets (p_1, p_2, \ldots, p_K)$
      \item[\textbf{Output:}]
      \Statex  $\mathbf{X}_{cached}$
    \end{algorithmic}
\end{algorithm}

\begin{algorithm}[!ht]
    \caption{Inference}
    \label{al:icarl-inference}
    \begin{algorithmic}[1]
        \item[\textbf{Input:}]
        \begin{itemize}
            \Statex \item $x \in \mathbb{R}^{C \times L \times H \times W}$ \Comment \textit{Testing video clips}
        \end{itemize}
        \item[\textbf{Require:}]
            \Statex Trained model $\mathcal{M}$ and weights $\Theta_{1\sim t-1}$
            \Statex $P=K\times n$ cached image set for all $n$ classes $\mathbf{X}=\left\{\mathbf{x}_1, \ldots, \mathbf{x}_n \right\}$, $\mathbf{x}_n \in \mathbb{R}^{K \times C \times L \times H \times W}$ \Comment \textit{Cached exemplar set}
        \item[\textbf{Compute exemplar feature means:}]
            \For{$k = 1:n$} 
                 \State $\mu_k \gets \frac{1}{K}\sum_{x_i\in \mathbf{x}_k} \mathcal{M}(x_i, \Theta_{1\sim t-1})$
            \EndFor
        \item[\textbf{Output:}]
            \Statex $y \gets \text{argmin}_{k=1,\ldots, n} \|\mathcal{M}(x, \Theta_{1\sim t-1}) - \mu_k \|$
    \end{algorithmic}
\end{algorithm}

Two evaluation metrics are used to validate the performance for each model during class incremental learning, which are mean accuracy and backward transfer (BWT)~\cite{she2019openloris}. Table~\ref{tab:train-test-matrix} shows an accuracy matrix $R$, which is able to observe the performance of a trained model changing over time. The row represents the model $\mathcal{M}_i$ trained on task $i$. The column represents the testing data from task $i$. The gray part is the BWT score, which measures the accuracy
over previously encountered tasks (average of gray elements in Table~\ref{tab:train-test-matrix}). BWT indicates the performance related to the memorization capability. The mean accuracy, average of last row elements in Table~\ref{tab:train-test-matrix}, demonstrates the overall performance on each task for the final model.   
\begin{table}[!ht]
\centering
\caption{Accuracy matrix $R$ during lifelong training, where $\mathcal{M}_i$ is the model trained using training data $Tr_i$ in task $T_i$, $Te_i$ is the testing data in task $T_i$, and $R_{ij} = $ classification accuracy of the model $\mathcal{M}_i$ training on $Tr_{i}$ and testing on $Te_{j}$. The number of tasks is $N$. Gray color represents the BWT score.} 
\begin{tabular}{c|cccc}
\toprule
 $R$   & $Te_{1}$ & $Te_{2}$ & $\cdots$ & $Te_{N}$ \\ \hline
$\mathcal{M}_{1}$ & $R_{11}$ & $R_{12}$ & $\cdots$ & $R_{1N}$ \\
$\mathcal{M}_{2}$ & \cellcolor{Gray}$R_{21}$ & $R_{22}$ & $\cdots$ &  $R_{2N}$ \\
$\cdots$ & \cellcolor{Gray}$\cdots$ & \cellcolor{Gray}$\cdots$ & $\cdots$ &  $\cdots$ \\ 
$\mathcal{M}_{N}$ & \cellcolor{Gray} $R_{N1}$ & \cellcolor{Gray}$R_{N2}$ & \cellcolor{Gray}$\cdots$  & $R_{NN}$ \\ \bottomrule
\end{tabular}
\label{tab:train-test-matrix}
\end{table}

\section{Experiments}
In this section, we describe experimental evaluation in this work. All models are tested on a public egocentric gesture dataset -- EgoGesture. Details of network settings are provided in the Appendix.
\subsection{EgoGesture Dataset}
EgoGesture is a recent multimodal large-scale video dataset for egocentric hand gesture recognition~\cite{zhang2018egogesture}. There are 83 classes of static and dynamic gestures collected from 6 diverse indoor and outdoor scenes. There are 24,161 video gesture samples and 2,953,224 frames, which are collected in RGB and depth modalities from 50 distinct participants. We follow the previous work~\cite{zhang2018egogesture,kopuklu2019real} to process the data, in which data was split by participants into training (60\%), validation (20\%) and testing (20\%). Participant IDs 2, 9, 11, 14, 18, 19, 28, 31, 41, 47 were used for testing and 1, 7, 12, 13, 24, 29, 33, 34, 35, 37 were used for validation. The rest of data was used for training. Similar to~\cite{kopuklu2019real}, we also included validation data during training. 

\subsection{Class Incremental Learning}
We focus on one of the lifelong learning scenarios in this study -- class incremental learning. Every time we extended the model, we added 5 new classes.
In order to get good generalization for our model on class incremental learning, we firstly trained our model using the first 40 classes (we will refer this initial task as task 0 in the later part of this paper). We then trained our model using 5 new classes (41 -- 45) as task 1. We repeated this procedure until task 9, in which data of classes 81 -- 85 was used. As a result, we have 10 tasks (including the initial training on the first 40 classes) over the class incremental learning process. 

\subsection{Implementation Details}
Three models were investigated in this work, which are ResNeXt-101-32f, ResNeXt-101-16f, ResNet-50-16f. Each model was tested by using 4 feature representations, which are depth input, RGB input, RGB and depth input (RGB-D) and two-stream. All models were first pretrained on Kinetics~\cite{kay2017kinetics,hara2018can}. 

Following previous work~\cite{kopuklu2019real}, we used the following methods to pre-process the data during training: (1) Each frame was firstly spatially resized to 112 $\times$ 112 pixels; (2) Each frame was scaled randomly with one of $\{1, \frac{1}{2^{1/4}}, \frac{1}{2^{3/4}}, \frac{1}{2}\}$ scales and then randomly cropped to size 112 $\times$ 112; (3) Spatial elastic displacement~\cite{simard2003best} with $\alpha = 1$ and $\sigma = 2$ was applied to cropped and scaled frames; (4) A fixed length clip (16-frame and 32-frame used in this work) was generated around the selected temporal position. If the video is shorter than the fixed length, we looped it as many times as possible; (5) We performed mean subtraction for each input channel, where mean values of ActivityNet~\cite{caba2015activitynet} were used. Finally we get the following types of inputs to our model: (1) Depth input, which has the size of 1 channel $\times$ 16/32 frames $\times$ 112 pixels $\times$ 112 pixels; (2) RGB input, which has the size of 3 channels $\times$ 16/32 frames $\times$ 112 pixels $\times$ 112 pixels; (3) RGB-D input, which has the size of 4 channels $\times$ 16/32 frames $\times$ 112 pixels $\times$ 112 pixels. Stochastic gradient descent was carried out to optimize the model when using backpropagation, which has a weight decay of $0.001$ and 0.9 for momentum. For task 0 training (first 40 classes), the learning rate was started from $0.001$ and divided by 10 at the 25th epoch. Training was completed after 50 epochs. For class incremental learning, the learning rate was started from $0.001$ and divided by 10 at the 6th epoch. Training was completed after 12 epochs. Batch size was set to 64 in the experiment.

During the testing session, testing frames were first scaled to the size of 112 $\times$ 112 and then cropped around a central position at scale 1. A testing video clip (with length 16 or 32) was generated at the central temporal position of a whole video. If the testing video clip was shorter than the required length, we looped it as many times as necessary. All testing frames were mean centered the same way those used for training.

\section{Results}
The presentation of results is divided into two parts. The first part compares the performance of different feature representations i.e., depth, RGB, RGB-D and two-stream. The second part compares the performance of different 3D ConvNets i.e., ResNeXt-101-32f, ResNeXt-101-16f and ResNet-50-16f. We use the joint training model as an upper bound comparison, which is trained by using the data of all classes. Mean accuracy and memorization capability are utilized to measure the performance.

\subsection{Comparison of Feature Representations}
\subsubsection{Final Model Accuracy for All Tasks}
Table~\ref{tab:inputs-compare} shows the mean accuracy across different tasks using different feature representations. It can be noticed that the two-stream approach achieves the highest accuracy for both joint training and class incremental training for all three different architectures, which indicates that two independent feature extractors for depth and RGB inputs should be beneficial for both joint training and lifelong learning. Previous work~\cite{abavisani2019improving,cao2017egocentric,zhang2018egogesture} has demonstrated that using RGB-D can outperform those only using one modality input in terms of joint training. However, it seems that the RGB-D feature representation is not beneficial to lifelong learning as it can be noticed that the depth feature representation performs better than the RGB-D feature representation for ResNext-101-16f and ResNet-50-16f during training the CatNet.       
\begin{table}[!htbp]
    \centering
    \caption{Mean accuracy for different feature representations. Bold text indicates the highest accuracy}
    \begin{tabular}{c|c|c|c}
        \toprule
        \multicolumn{2}{c|}{Method} & {Joint training} & {CatNet}\\ \hline
        \multirow{4}{*}{ResNeXt-101-32f}& {Depth} & {0.909} & {0.845}\\ 
        &{RGB} & {0.905} & {0.859}\\ 
        &{RGB-D} & {0.922} & {0.861}\\ 
        &{Two-stream} & {\textbf{0.932}} & {\textbf{0.884}}\\ \hline
        
        \multirow{4}{*}{ResNeXt-101-16f}& {Depth} & {0.883} & {0.840}\\ 
        &{RGB} & {0.850} & {0.826}\\ 
        &{RGB-D} & {0.891} & {0.834}\\ 
        &{Two-stream} & {\textbf{0.907}} & {\textbf{0.865}}\\ \hline
        
        \multirow{4}{*}{ResNet-50-16f}& {Depth} & {0.870} & {0.843}\\
        &{RGB} & {0.865} & {0.792}\\ 
        &{RGB-D} & {0.867} & {0.830}\\ 
        &{Two-stream} & {\textbf{0.900}} & {\textbf{0.854}}\\ \bottomrule
    \end{tabular}
    \label{tab:inputs-compare}
\end{table}

\subsubsection{Memorization Capability}
\begin{figure*}[!ht]
     \begin{minipage}{.72\textwidth}
     \subfigure[Depth]{\includegraphics[width=.21\textwidth]{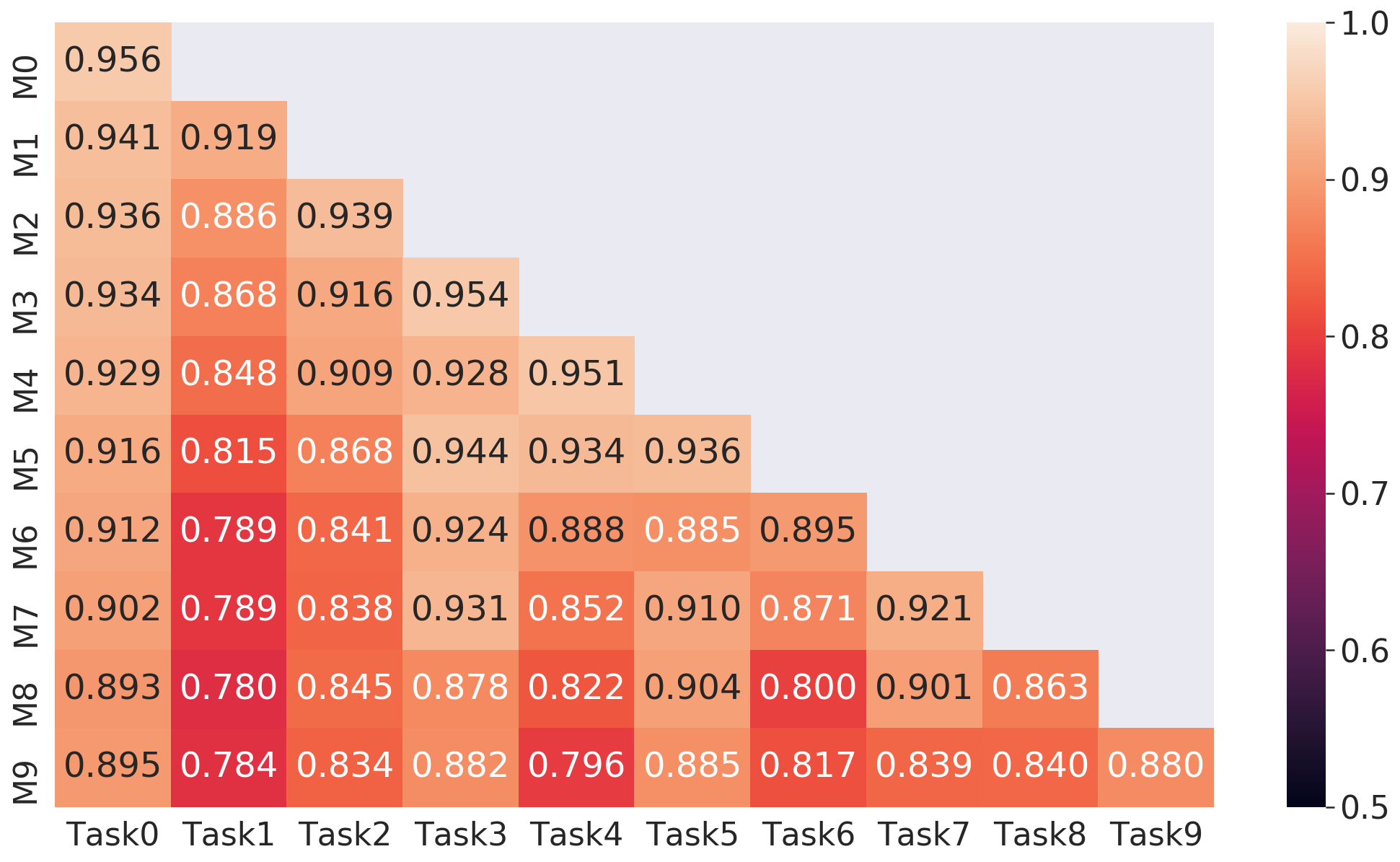} }
     \subfigure[RGB]{\includegraphics[width=.21\textwidth]{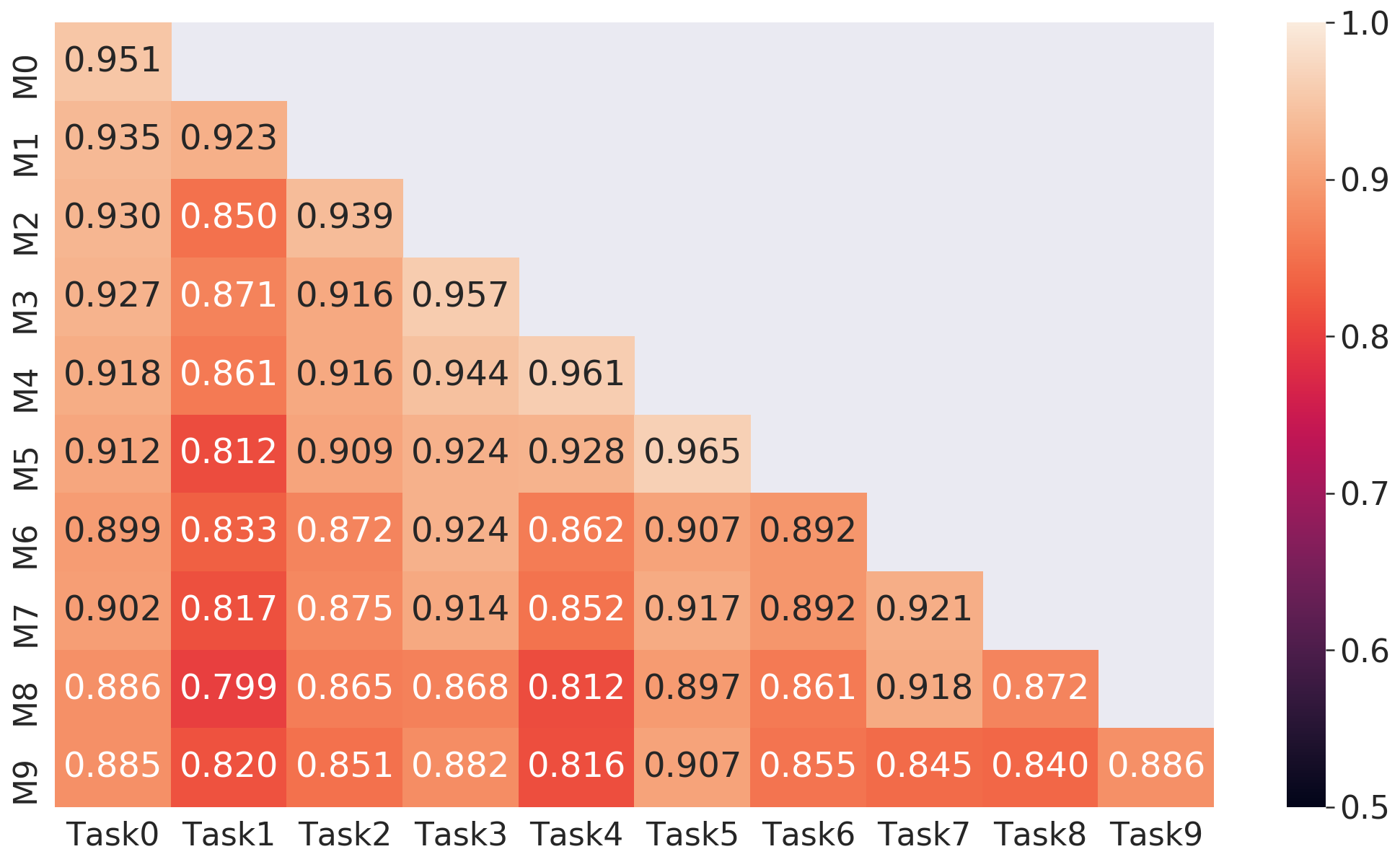} }
     \subfigure[RGB-D]{\includegraphics[width=.21\textwidth]{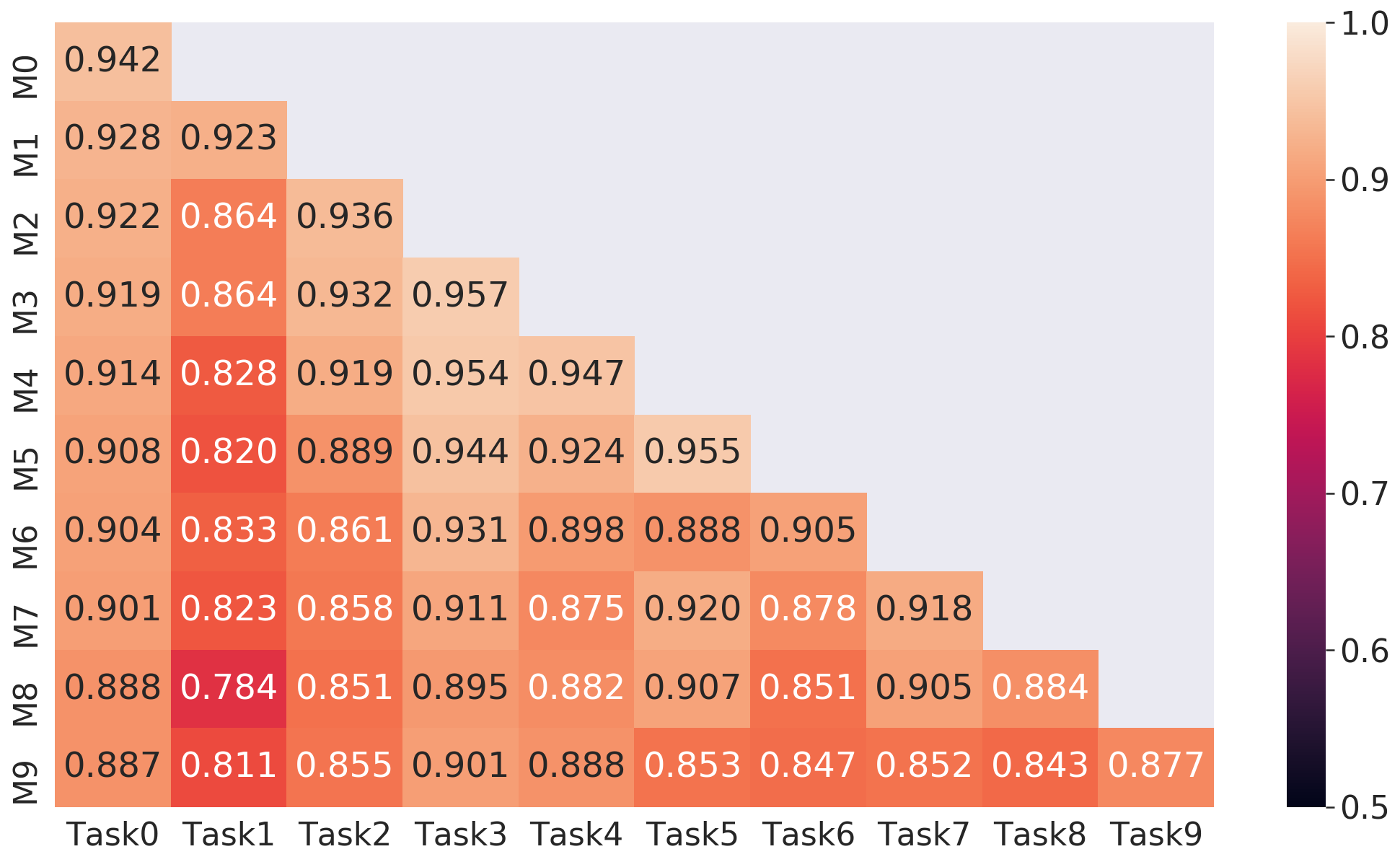} }
     \subfigure[Two-stream]{\includegraphics[width=.21\textwidth]{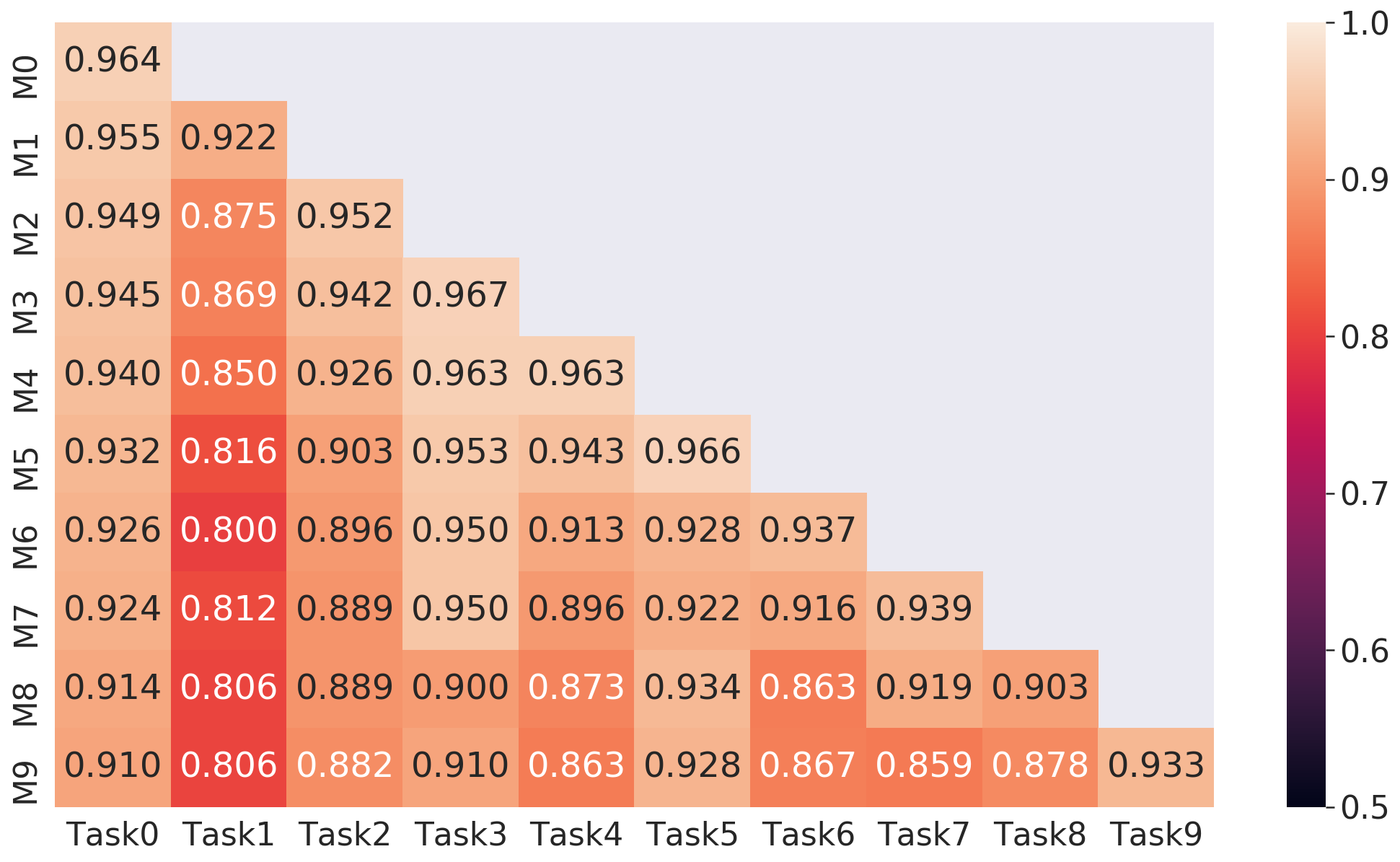}}
    \end{minipage}
    \vspace{-75pt}
    
    \begin{minipage}{.72\textwidth} 
    \subfigure[Depth]{\includegraphics[width=.21\textwidth]{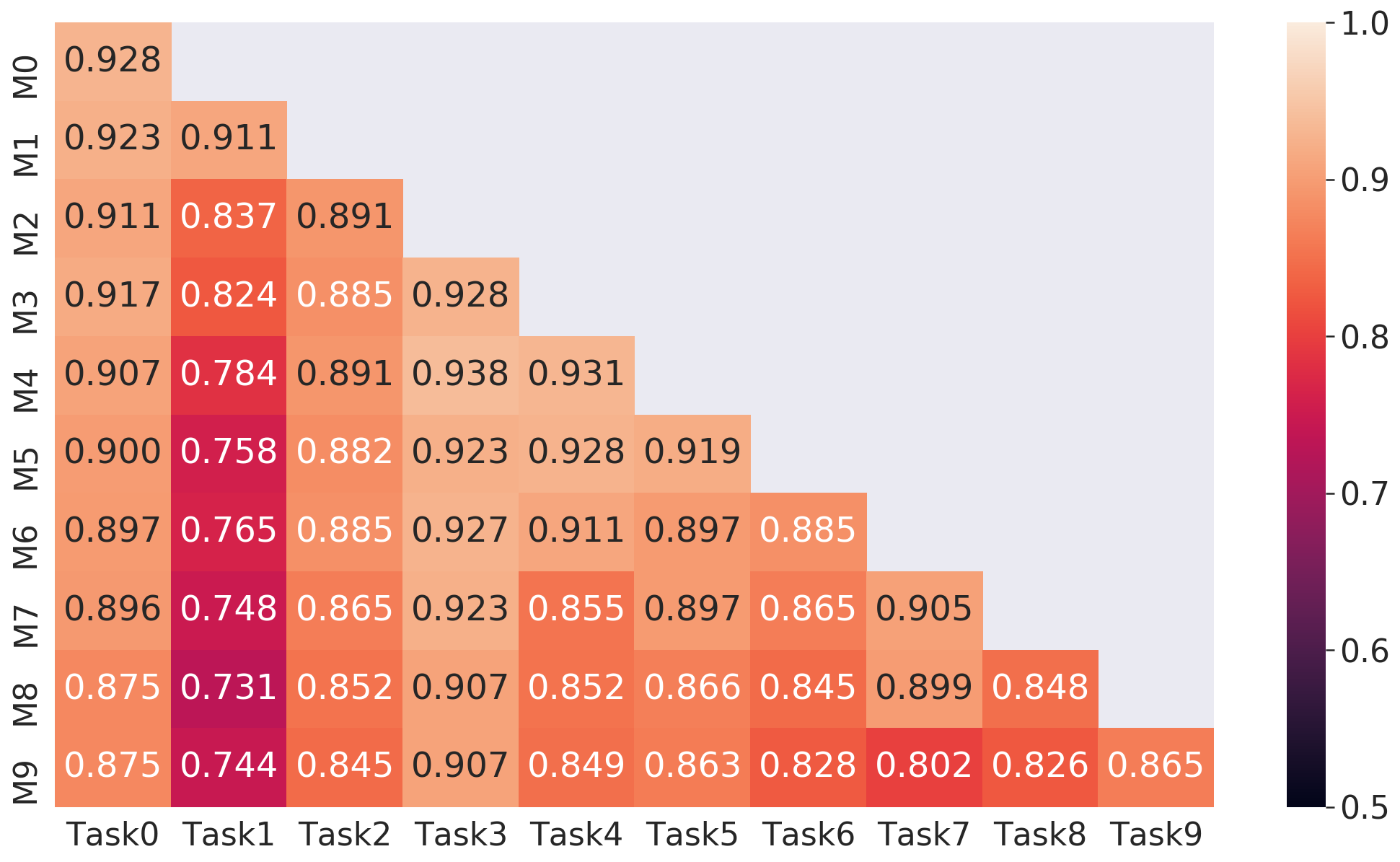} }
     \subfigure[RGB]{\includegraphics[width=.21\textwidth]{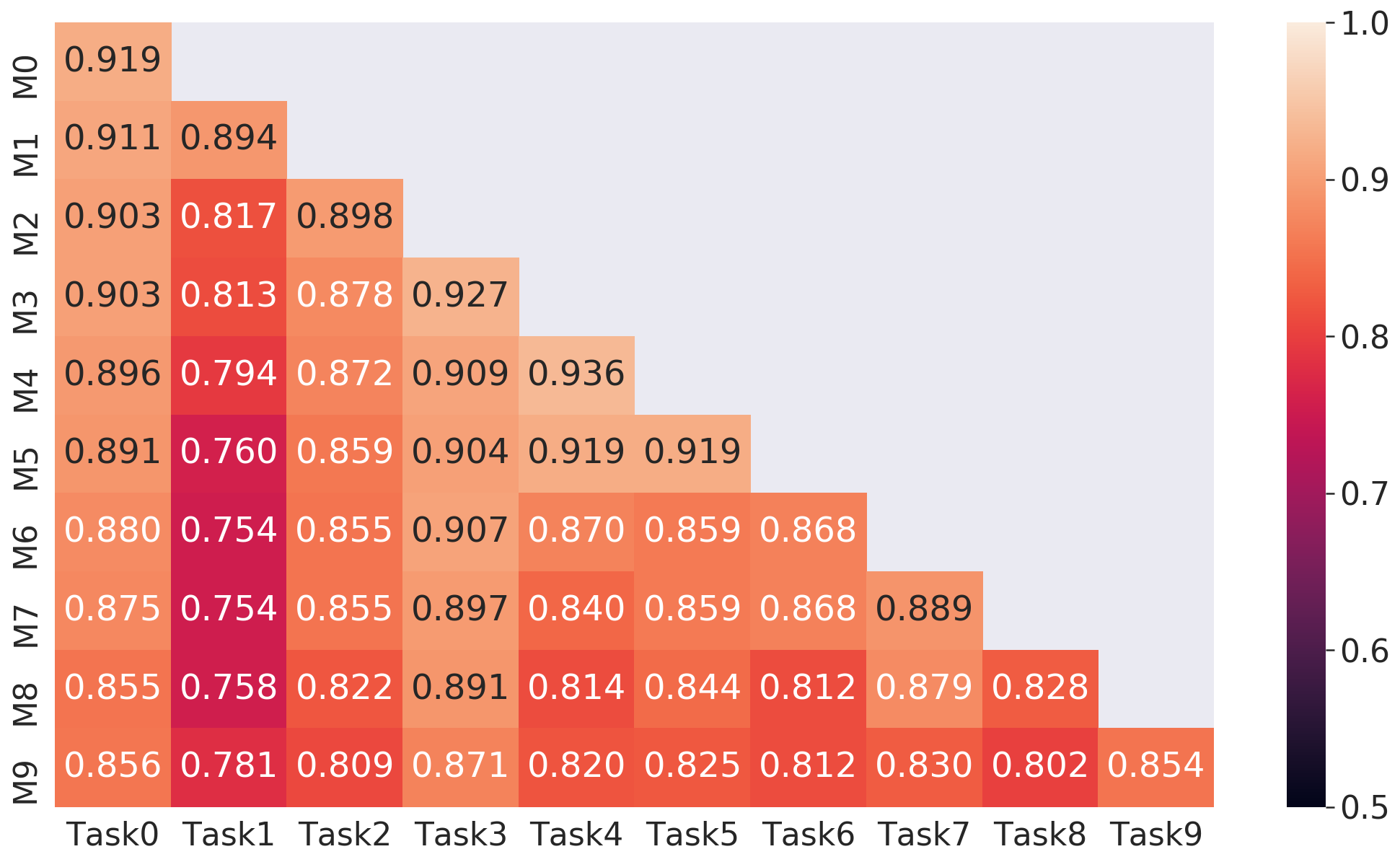} }
     \subfigure[RGB-D]{\includegraphics[width=.21\textwidth]{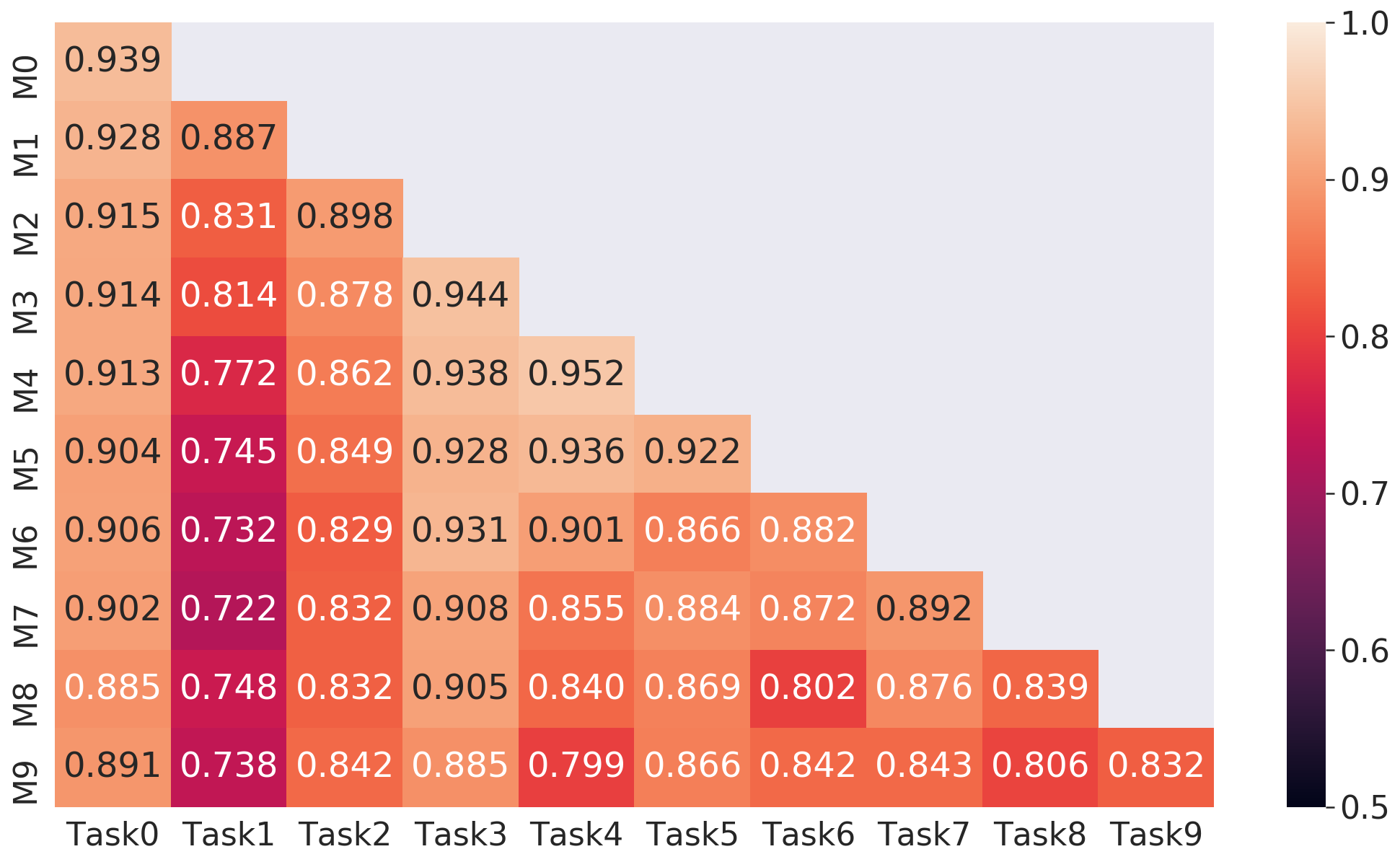} }
    \subfigure[Two-stream]{\includegraphics[width=.21\textwidth]{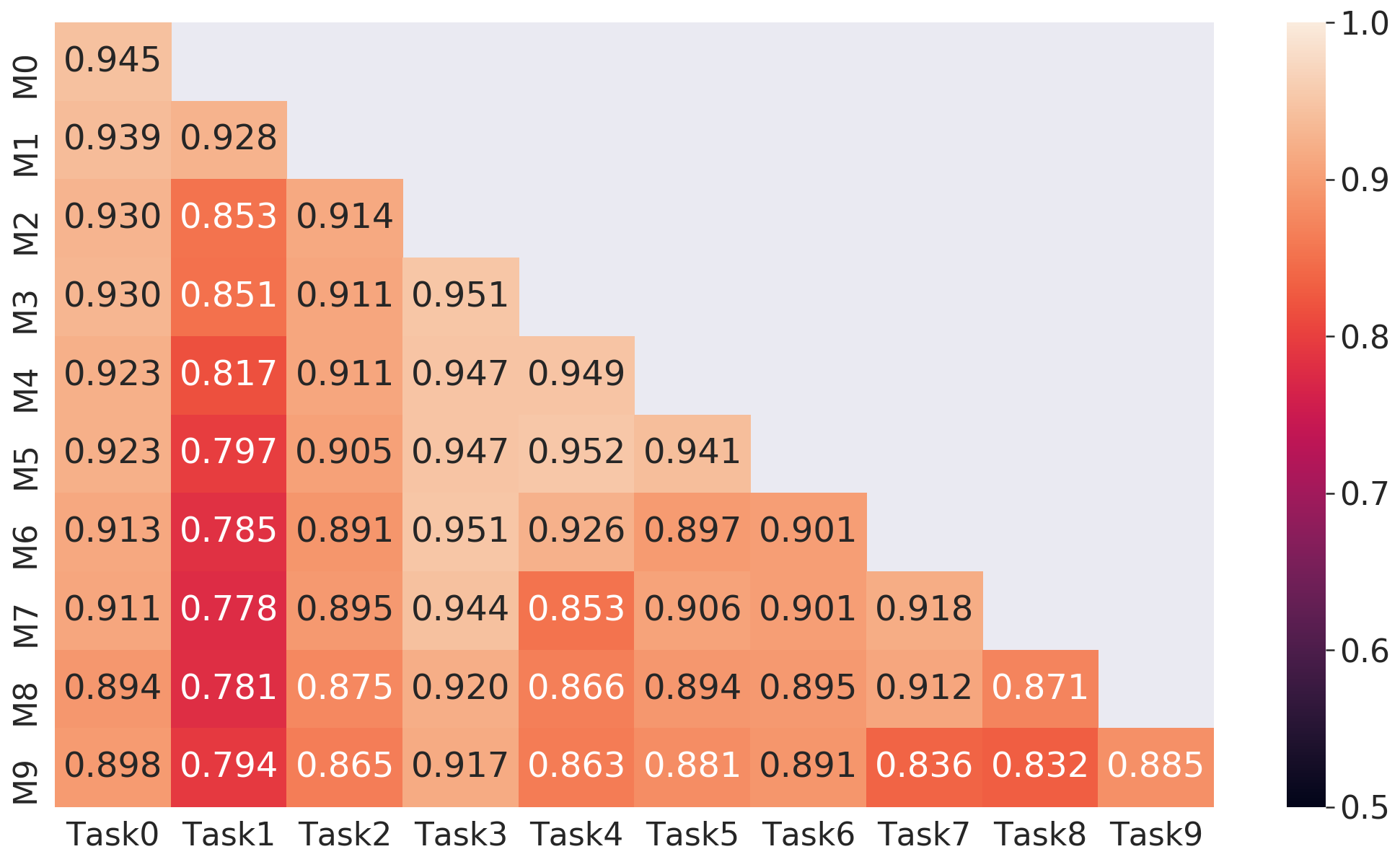} }
    \end{minipage}
    \vspace{-75pt}
    \hspace{-45pt}
    \begin{minipage}{.35\textwidth}
    \captionof{table}{Memorization capability across different feature representations. Bold text indicates the highest accuracy.}
    \label{tab:inputs-BWT}
    \begin{tabular}{c|c|c}
        \toprule
        \multicolumn{2}{c|}{Model} & {BWT}\\ \hline
        \multirow{4}{*}{ResNeXt-101-32f}& {Depth} & {0.873}\\ 
        &{RGB} & {0.880}\\ 
        &{RGB-D}& {0.882}\\ 
        &{Two-stream} & {\textbf{0.900}}\\ \hline
        
        \multirow{4}{*}{ResNeXt-101-16f}& {Depth}& {0.865}\\ 
        &{RGB} & {0.849}\\ 
        &{RGB-D} & {0.856}\\ 
        &{Two-stream} & {\textbf{0.887}}\\ \hline
        
        \multirow{4}{*}{ResNet-50-16f}& {Depth} & {0.863}\\
        &{RGB} & {0.823}\\ 
        &{RGB-D} & {0.853}\\ 
        &{Two-stream}& {\textbf{0.880}}\\ \bottomrule
    \end{tabular}
    \end{minipage}
    
    \begin{minipage}{.72\textwidth}
    \subfigure[Depth]{\includegraphics[width=.21\textwidth]{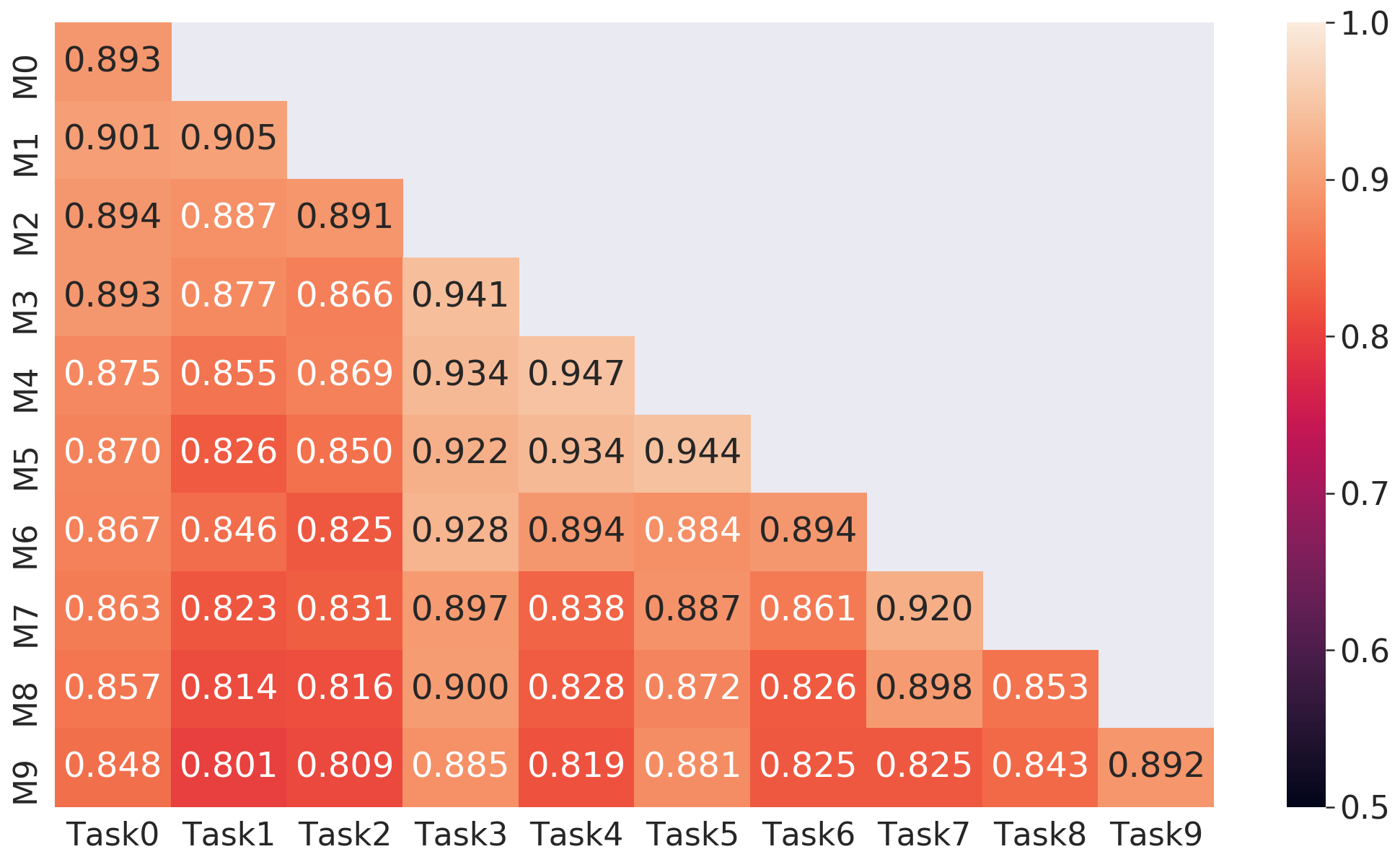} }
     \subfigure[RGB]{\includegraphics[width=.21\textwidth]{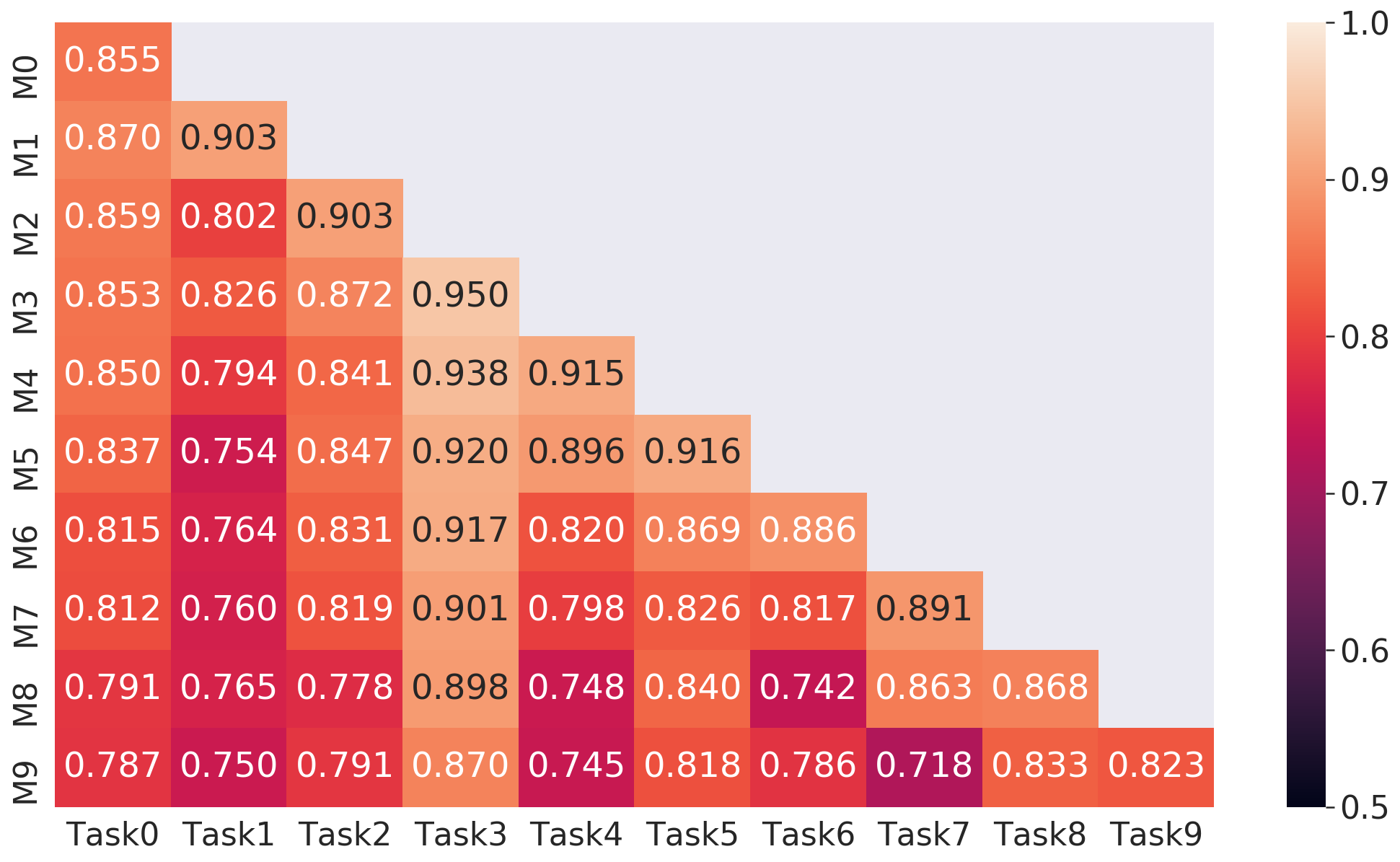} }
     \subfigure[RGB-D]{\includegraphics[width=.2\textwidth]{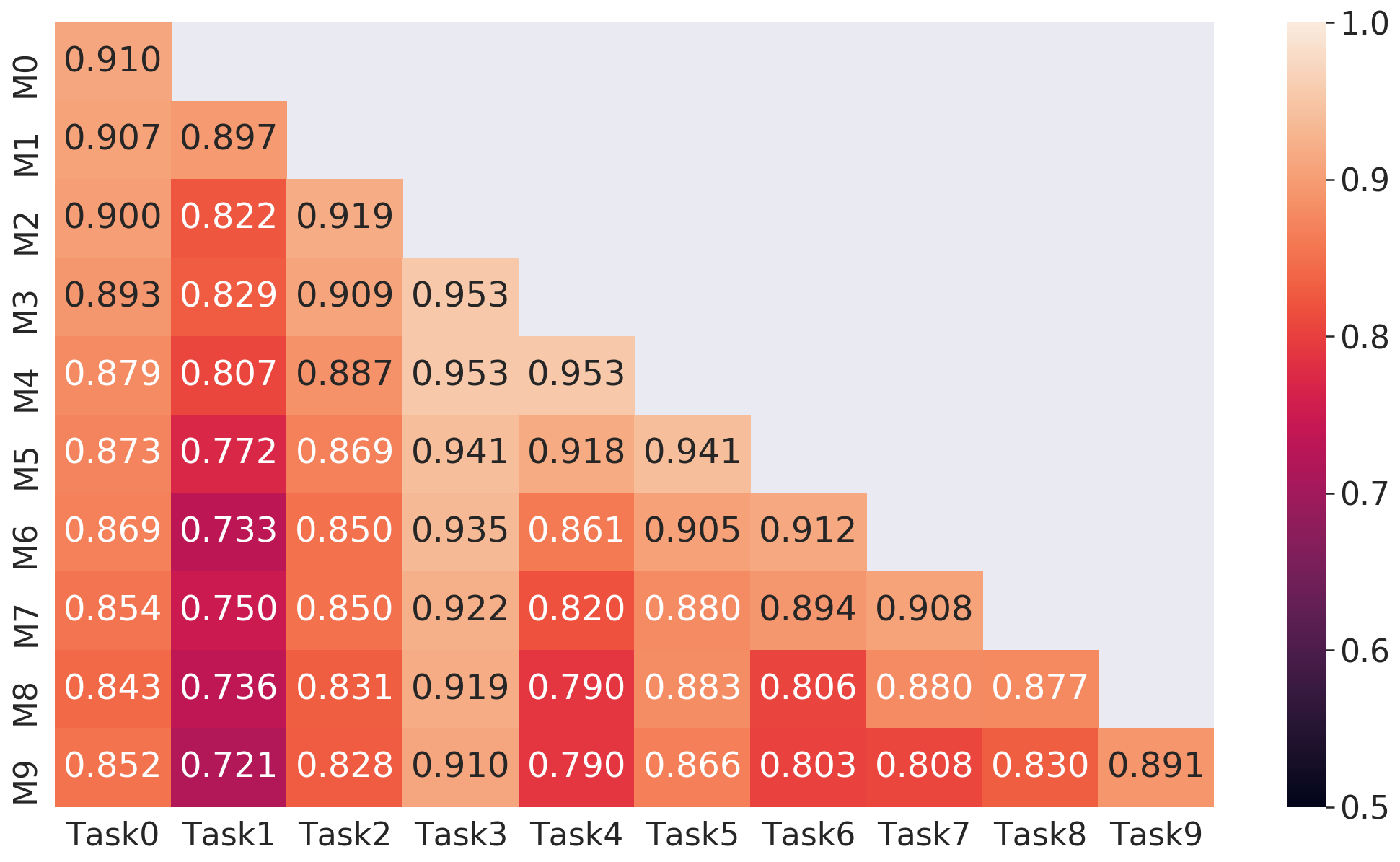} }
    \subfigure[Two-stream]{\includegraphics[width=.21\textwidth]{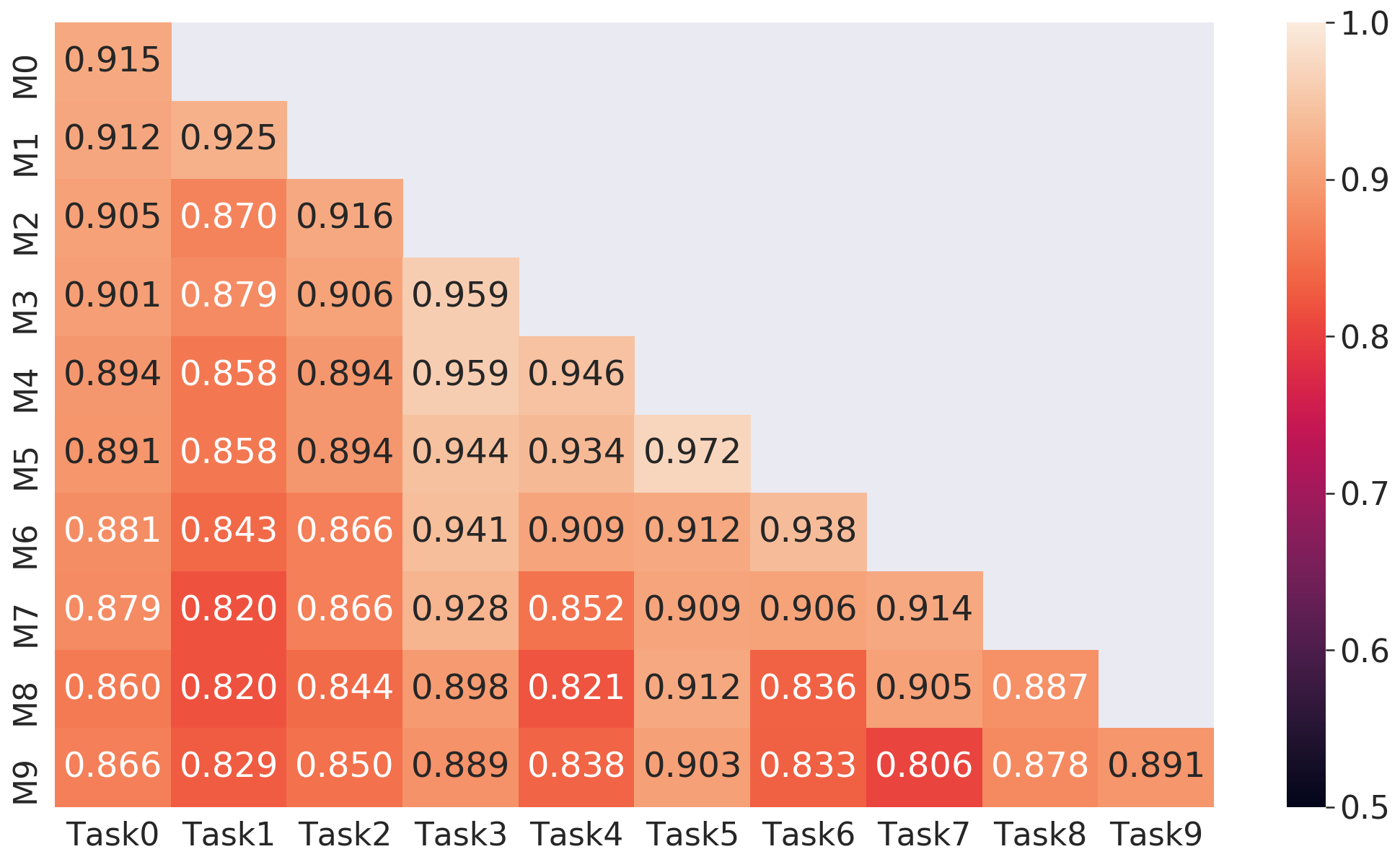} }
    \end{minipage}
    \caption{Left: Classification accuracy matrix $R$ for three architectures i.e., ResNext-101-32f (top), ResNext-101-16f (middle) and ResNet-50-16f (bottom). The vertical axis is the model $\mathcal{M}_i$ trained on the task $T_i$. The horizontal axis is the task $T_i$ data. Lighter color indicates better performance. Right: Table summary of the figure on the left.}
    \label{fig-CM}
\end{figure*}

BWT is carried out in this work for measuring the memorization capability. Table~\ref{tab:inputs-BWT} summarizes that BWT in Figure~\ref{fig-CM} (left). Compared to other feature representation approaches, the two-stream feature representation shows that the model produces lighter color in the matrix over time, which indicates a better memorization capability. Similar to the mean accuracy, the RGB-D feature representation performs worse than the depth feature representation for ResNeXt-101-16f and ResNet-50-16f. These results indicate that the one-stream CatNet is not able to fully make use of RGB-D information when only concatenating RGB and depth as an input to the model with respect to EgoGesture video recognition. Thus we provide such a two-stream strategy which shows good performance for both joint training and lifelong learning.   

\subsection{Comparison of Architectures}
\subsubsection{Final Model Accuracy for All Tasks}
Table~\ref{tab:architecture-compare} shows the mean accuracy of joint training and class incremental training across different architectures. It can be noticed that ResNeXt-101-32f achieves the best performance for both joint training and lifelong learning across different feature representations. ResNeXt-101-32f has the same depth as ResNeXt-101-16f, but a longer temporal frame length is used for the input clip when using ResNeXt-101-32f, which is able to preserve more temporal information from videos. 
\begin{table}[!htbp]
    \caption{Comparison between architectures using mean accuracy. Bold text indicates the best performance.}
    \centering
    \begin{tabular}{c|c|c|c}
        \toprule
        \multicolumn{2}{c|}{Method} & {Joint training} & {CatNet}\\ \hline
        \multirow{3}{*}{Depth}& {ResNeXt-101-32f} & {\textbf{0.909}} & {\textbf{0.845}}\\ 
        &{ResNeXt-101-16f} & {0.883} & {0.840} \\ 
        &{ResNet-50-16f} & {0.870} & {0.843}\\ \hline
        
        \multirow{3}{*}{RGB}& {ResNeXt-101-32f} & {\textbf{0.905}} & {\textbf{0.859}}\\ 
        &{ResNeXt-101-16f} & {0.850} & {0.826}\\ 
        &{ResNet-50-16f} & {0.865} & {0.792}\\ \hline
        
        \multirow{3}{*}{RGB-D}& {ResNeXt-101-32f} & {\textbf{0.922}} & {\textbf{0.861}}\\ 
        &{ResNeXt-101-16f} & {0.891} & {0.834}\\ 
        &{ResNet-50-16f} & {0.867} & {0.830}\\ \hline
        
        \multirow{3}{*}{Two-stream}& {ResNeXt-101-32f} & {\textbf{0.932}} & {\textbf{0.884}}\\ 
        &{ResNeXt-101-16f} & {0.907} & {0.865}\\ 
        &{ResNet-50-16f} & {0.900} & {0.854}\\ \bottomrule
    \end{tabular}
    \label{tab:architecture-compare}
\end{table}

\subsubsection{Memorization Capability}
Table~\ref{tab:architecture-BWT} shows BWT across three different architectures for each feature representation. The rank of BWT for each feature representation is the same, which is ResNeXt-101-32f, ResNeXt-101-16f and ResNet-50-16f from high to low. Because BWT can also be affected by the initial performance of the model i.e., initial classification performance for the first 40 classes in our case, we also test the initial classification accuracy for the first 40 classes for each model as seen in Table~\ref{tab:architecture-BWT}. The initial accuracy has exactly the same order as BWT in terms of ranking the three architectures for each representation i.e., ResNeXt-101-32f, ResNeXt-101-16f and ResNet-50-16f. This indicates that a deeper model is able to improve the initial performance on the initial task (task 0) but can not benefit the memorization capability on a lifelong learning task.  
\begin{table}[!htbp]
    \caption{Comparison between different architectures for memorization capability. Bold text indicates the best performance.}
    \centering
    \begin{tabular}{c|c|c|c}
        \toprule
        \multicolumn{2}{c|}{Model} & {BWT} & {Initial accuracy}\\ \hline
        \multirow{4}{*}{Depth}& {ResNeXt-101-32f} & {\textbf{0.873}} & {\textbf{0.956}}\\ 
        &{ResNeXt-101-16f} & {0.865} & {0.928}\\ 
        &{ResNet-50-16f} & {0.863} & {0.894}\\ \hline
        
        \multirow{4}{*}{RGB}& {ResNeXt-101-32f}& {\textbf{0.880}} & {\textbf{0.951}}\\ 
        &{ResNeXt-101-16f} & {0.849} & {0.919}\\ 
        &{ResNet-50-16f} & {0.823} & {0.878}\\ \hline
        
        \multirow{4}{*}{RGB-D}& {ResNeXt-101-32f}& {\textbf{0.882}} & {\textbf{0.942}}\\ 
        &{ResNeXt-101-16f} & {0.856} & {0.939}\\ 
        &{ResNet-50-16f} & {0.853} & {0.910}\\ \hline
        
        \multirow{4}{*}{Two-stream}& {ResNeXt-101-32f} & {\textbf{0.900}} & {\textbf{0.964}}\\
        &{ResNeXt-101-16f} & {0.887} & {0.945}\\ 
        &{ResNet-50-16f}& {0.880} & {0.915}\\ \bottomrule
    \end{tabular}
    \label{tab:architecture-BWT}
\end{table}


\section{Discussion}
\begin{figure}[!ht]
    \centering
    \includegraphics[width=.4\textwidth]{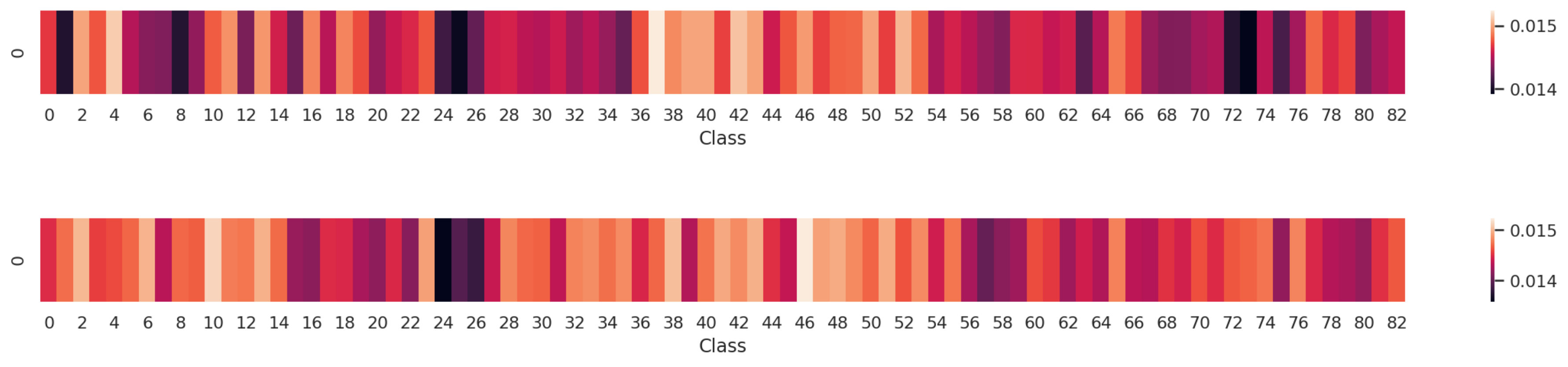}
    \caption{Visualization of features extracted by the two-stream CatNet.}
    \label{fig:feature_viz}
\end{figure}

Figure~\ref{fig:feature_viz} shows the features produced by a feature mean matrix according to the depth input and the RGB input respectively for all 83 classes (horizontal represents different classes). Given a feature mean matrix $\mathcal{S} \in \mathbb{R}^{m \times c}$, where $m$ is the number of features and $c$ is the number of classes, it is not easy to visualize because of the large feature number. We average $\mathcal{S}$ over the feature dimension, which is derived as $\frac{1}{m}\sum^{m}_i \mathcal{S}_i$, for visualization. It can be seen that the depth feature representation and the RGB feature representation are quite different from each other. Because our model uses the mean exemplar set as a reference for classification, the two-stream approach, which fuses depth features and RGB features from the second last layer, can be beneficial for this case. 


\vspace{-20pt}
\section{Conclusion}

In this paper, we investigate class incremental learning in the context of egocentric gesture video recognition, in which we address the issue in such scenarios for real-wold applications is that may easily become necessary to add new gestures to the system. A 3D convolution based framework named CatNet is introduced and we demonstrate the efficacy of CatNets on the EgoGesture dataset, in which the performance on the class incremental task does not drop significantly compared to joint training. Importantly, we propose the use of a two-stream architecture for the CatNet, in which two 3D ConvNets are trained independently by feeding RGB and depth inputs. Results demonstrate that the two-stream CatNet performs better than 3 other one-stream CatNets both on the mean accuracy and the memorization capability. Results also demonstrate that CatNet exhibits some forgetting of knowledge, which can be further investigated in the future.


{\small
\bibliographystyle{ieee_fullname}
\bibliography{egbib}

\begin{thebibliography}{10}\itemsep=-1pt

\bibitem{abavisani2019improving}
Mahdi Abavisani, Hamid Reza~Vaezi Joze, and Vishal~M Patel.
\newblock Improving the performance of unimodal dynamic hand-gesture
  recognition with multimodal training.
\newblock In {\em Proceedings of the IEEE Conference on Computer Vision and
  Pattern Recognition}, pages 1165--1174, 2019.

\bibitem{abu2016youtube}
Sami Abu-El-Haija, Nisarg Kothari, Joonseok Lee, Paul Natsev, George Toderici,
  Balakrishnan Varadarajan, and Sudheendra Vijayanarasimhan.
\newblock You{T}ube-8{M}: {A} large-scale video classification benchmark.
\newblock {\em arXiv preprint arXiv:1609.08675}, 2016.

\bibitem{benna2016computational}
Marcus~K Benna and Stefano Fusi.
\newblock Computational principles of synaptic memory consolidation.
\newblock {\em Nature neuroscience}, 19(12):1697, 2016.

\bibitem{caba2015activitynet}
Fabian Caba~Heilbron, Victor Escorcia, Bernard Ghanem, and Juan Carlos~Niebles.
\newblock Activitynet: {A} large-scale video benchmark for human activity
  understanding.
\newblock In {\em Proceedings of the IEEE conference on computer vision and
  pattern recognition}, pages 961--970, 2015.

\bibitem{cao2017egocentric}
Congqi Cao, Yifan Zhang, Yi Wu, Hanqing Lu, and Jian Cheng.
\newblock Egocentric gesture recognition using recurrent 3{D} convolutional
  neural networks with spatiotemporal transformer modules.
\newblock In {\em Proceedings of the IEEE International Conference on Computer
  Vision}, pages 3763--3771, 2017.

\bibitem{carreira2017quo}
Joao Carreira and Andrew Zisserman.
\newblock Quo vadis, action recognition? {A} new model and the kinetics
  dataset.
\newblock In {\em proceedings of the IEEE Conference on Computer Vision and
  Pattern Recognition}, pages 6299--6308, 2017.

\bibitem{tejo2018}
Tejo Chalasani, Jan Ondrej, and Aljosa Smolic.
\newblock Egocentric gesture recognition for head mounted ar devices.
\newblock {\em Adjunct Proceedings of the IEEE and ACM International Symposium
  for Mixed and Augmented Reality}, 2018.

\bibitem{tejo2019}
Tejo Chalasani and Aljosa Smolic.
\newblock Simultaneous segmentation and recognition: Towards more accurate ego
  gesture recognition.
\newblock 2019.

\bibitem{diba2017temporal}
Ali Diba, Mohsen Fayyaz, Vivek Sharma, Amir~Hossein Karami, Mohammad~Mahdi
  Arzani, Rahman Yousefzadeh, and Luc Van~Gool.
\newblock Temporal 3{D} convnets: {N}ew architecture and transfer learning for
  video classification.
\newblock {\em arXiv preprint arXiv:1711.08200}, 2017.

\bibitem{dosovitskiy2015flownet}
Alexey Dosovitskiy, Philipp Fischer, Eddy Ilg, Philip Hausser, Caner Hazirbas,
  Vladimir Golkov, Patrick Van Der~Smagt, Daniel Cremers, and Thomas Brox.
\newblock Flow{N}et: {L}earning optical flow with convolutional networks.
\newblock In {\em Proceedings of the IEEE international conference on computer
  vision}, pages 2758--2766, 2015.

\bibitem{feichtenhofer2016convolutional}
Christoph Feichtenhofer, Axel Pinz, and Andrew Zisserman.
\newblock Convolutional two-stream network fusion for video action recognition.
\newblock In {\em Proceedings of the IEEE conference on computer vision and
  pattern recognition}, pages 1933--1941, 2016.

\bibitem{feng2019challenges}
Fan Feng, Rosa~HM Chan, Xuesong Shi, Yimin Zhang, and Qi She.
\newblock Challenges in task incremental learning for assistive robotics.
\newblock {\em IEEE Access}, 2019.

\bibitem{fernando2017pathnet}
Chrisantha Fernando, Dylan Banarse, Charles Blundell, Yori Zwols, David Ha,
  Andrei~A Rusu, Alexander Pritzel, and Daan Wierstra.
\newblock {Pathnet: Evolution channels gradient descent in super neural
  networks}.
\newblock {\em arXiv preprint arXiv:1701.08734}, 2017.

\bibitem{fusi2005cascade}
Stefano Fusi, Patrick~J Drew, and Larry~F Abbott.
\newblock Cascade models of synaptically stored memories.
\newblock {\em Neuron}, 45(4):599--611, 2005.

\bibitem{gepperth2016bio}
Alexander Gepperth and Cem Karaoguz.
\newblock A bio-inspired incremental learning architecture for applied
  perceptual problems.
\newblock {\em Cognitive Computation}, 8(5):924--934, 2016.

\bibitem{hara2018can}
Kensho Hara, Hirokatsu Kataoka, and Yutaka Satoh.
\newblock Can spatiotemporal {3D CNNs} retrace the history of {2D CNNs} and
  {ImageNet}?
\newblock In {\em Proceedings of the IEEE conference on Computer Vision and
  Pattern Recognition}, pages 6546--6555, 2018.

\bibitem{ji20123d}
Shuiwang Ji, Wei Xu, Ming Yang, and Kai Yu.
\newblock 3{D} convolutional neural networks for human action recognition.
\newblock {\em IEEE transactions on pattern analysis and machine intelligence},
  35(1):221--231, 2012.

\bibitem{karpathy2014large}
Andrej Karpathy, George Toderici, Sanketh Shetty, Thomas Leung, Rahul
  Sukthankar, and Li Fei-Fei.
\newblock Large-scale video classification with convolutional neural networks.
\newblock In {\em Proceedings of the IEEE conference on Computer Vision and
  Pattern Recognition}, pages 1725--1732, 2014.

\bibitem{kay2017kinetics}
Will Kay, Joao Carreira, Karen Simonyan, Brian Zhang, Chloe Hillier, Sudheendra
  Vijayanarasimhan, Fabio Viola, Tim Green, Trevor Back, Paul Natsev, et~al.
\newblock The kinetics human action video dataset.
\newblock {\em arXiv preprint arXiv:1705.06950}, 2017.

\bibitem{kemker2017fearnet}
Ronald Kemker and Christopher Kanan.
\newblock {Fearnet: Brain-inspired model for incremental learning}.
\newblock {\em arXiv preprint arXiv:1711.10563}, 2017.

\bibitem{kopuklu2019real}
Okan K{\"o}p{\"u}kl{\"u}, Ahmet Gunduz, Neslihan Kose, and Gerhard Rigoll.
\newblock Real-time hand gesture detection and classification using
  convolutional neural networks.
\newblock In {\em 2019 14th IEEE International Conference on Automatic Face \&
  Gesture Recognition (FG 2019)}, pages 1--8. IEEE, 2019.

\bibitem{krizhevsky2012imagenet}
Alex Krizhevsky, Ilya Sutskever, and Geoffrey~E Hinton.
\newblock {ImageNet} classification with deep convolutional neural networks.
\newblock In {\em Advances in neural information processing systems}, pages
  1097--1105, 2012.

\bibitem{liu2013learning}
Li Liu and Ling Shao.
\newblock Learning discriminative representations from {RGB-D} video data.
\newblock In {\em Twenty-Third International Joint Conference on Artificial
  Intelligence}, 2013.

\bibitem{long2015fully}
Jonathan Long, Evan Shelhamer, and Trevor Darrell.
\newblock Fully convolutional networks for semantic segmentation.
\newblock In {\em Proceedings of the IEEE conference on computer vision and
  pattern recognition}, pages 3431--3440, 2015.

\bibitem{mccloskey1989catastrophic}
Michael McCloskey and Neal~J Cohen.
\newblock Catastrophic interference in connectionist networks: The sequential
  learning problem.
\newblock In {\em Psychology of learning and motivation}, volume~24, pages
  109--165. Elsevier, 1989.

\bibitem{ohn2014hand}
Eshed Ohn-Bar and Mohan~Manubhai Trivedi.
\newblock Hand gesture recognition in real time for automotive interfaces: {A}
  multimodal vision-based approach and evaluations.
\newblock {\em IEEE transactions on intelligent transportation systems},
  15(6):2368--2377, 2014.

\bibitem{parisi2019continual}
German~I Parisi, Ronald Kemker, Jose~L Part, Christopher Kanan, and Stefan
  Wermter.
\newblock Continual lifelong learning with neural networks: {A} review.
\newblock {\em Neural Networks}, 2019.

\bibitem{qiu2017learning}
Zhaofan Qiu, Ting Yao, and Tao Mei.
\newblock Learning spatio-temporal representation with pseudo-3{D} residual
  networks.
\newblock In {\em proceedings of the IEEE International Conference on Computer
  Vision}, pages 5533--5541, 2017.

\bibitem{rebuffi2017icarl}
Sylvestre-Alvise Rebuffi, Alexander Kolesnikov, Georg Sperl, and Christoph~H
  Lampert.
\newblock icarl: Incremental classifier and representation learning.
\newblock In {\em Proceedings of the IEEE conference on Computer Vision and
  Pattern Recognition}, pages 2001--2010, 2017.

\bibitem{redmon2016you}
Joseph Redmon, Santosh Divvala, Ross Girshick, and Ali Farhadi.
\newblock You only look once: {U}nified, real-time object detection.
\newblock In {\em Proceedings of the IEEE conference on computer vision and
  pattern recognition}, pages 779--788, 2016.

\bibitem{she2019openloris}
Qi She, Fan Feng, Xinyue Hao, Qihan Yang, Chuanlin Lan, Vincenzo Lomonaco,
  Xuesong Shi, Zhengwei Wang, Yao Guo, Yimin Zhang, et~al.
\newblock {OpenLORIS-Object}: {A} dataset and benchmark towards lifelong object
  recognition.
\newblock {\em arXiv preprint arXiv:1911.06487}, 2019.

\bibitem{she2018reduced}
Qi She, Yuan Gao, Kai Xu, and Rosa~HM Chan.
\newblock Reduced-rank linear dynamical systems.
\newblock In {\em Thirty-Second AAAI Conference on Artificial Intelligence
  (AAAI)}, 2018.

\bibitem{she2019neural}
Qi She and Anqi Wu.
\newblock Neural dynamics discovery via gaussian process recurrent neural
  networks.
\newblock {\em arXiv preprint arXiv:1907.00650}, 2019.

\bibitem{simard2003best}
Patrice~Y Simard, David Steinkraus, John~C Platt, et~al.
\newblock Best practices for convolutional neural networks applied to visual
  document analysis.
\newblock In {\em Icdar}, volume~3, 2003.

\bibitem{simonyan2014two}
Karen Simonyan and Andrew Zisserman.
\newblock Two-stream convolutional networks for action recognition in videos.
\newblock In {\em Advances in neural information processing systems}, pages
  568--576, 2014.

\bibitem{soomro2012ucf101}
Khurram Soomro, Amir~Roshan Zamir, and Mubarak Shah.
\newblock {UCF101}: {A} dataset of 101 human actions classes from videos in the
  wild.
\newblock {\em arXiv preprint arXiv:1212.0402}, 2012.

\bibitem{tran2015learning}
Du Tran, Lubomir Bourdev, Rob Fergus, Lorenzo Torresani, and Manohar Paluri.
\newblock Learning spatiotemporal features with 3{D} convolutional networks.
\newblock In {\em Proceedings of the IEEE international conference on computer
  vision}, pages 4489--4497, 2015.

\bibitem{van2019three}
Gido~M van~de Ven and Andreas~S Tolias.
\newblock Three scenarios for continual learning.
\newblock {\em arXiv preprint arXiv:1904.07734}, 2019.

\bibitem{wan2016chalearn}
Jun Wan, Yibing Zhao, Shuai Zhou, Isabelle Guyon, Sergio Escalera, and Stan~Z
  Li.
\newblock Chalearn looking at people {RGB-D} isolated and continuous datasets
  for gesture recognition.
\newblock In {\em Proceedings of the IEEE Conference on Computer Vision and
  Pattern Recognition Workshops}, pages 56--64, 2016.

\bibitem{wang2016exploring}
Cheng Wang, Haojin Yang, and Christoph Meinel.
\newblock Exploring multimodal video representation for action recognition.
\newblock In {\em 2016 International Joint Conference on Neural Networks
  (IJCNN)}, pages 1924--1931. IEEE, 2016.

\bibitem{wang2016temporal}
Limin Wang, Yuanjun Xiong, Zhe Wang, Yu Qiao, Dahua Lin, Xiaoou Tang, and Luc
  Van~Gool.
\newblock Temporal segment networks: {T}owards good practices for deep action
  recognition.
\newblock In {\em European conference on computer vision}, pages 20--36.
  Springer, 2016.

\bibitem{wang2020use}
Zhengwei Wang, Graham Healy, Alan~F Smeaton, and Tomas~E Ward.
\newblock Use of neural signals to evaluate the quality of generative
  adversarial network performance in facial image generation.
\newblock {\em Cognitive Computation}, 12(1):13--24, 2020.

\bibitem{wang2019neuroscore}
Zhengwei Wang, Qi She, Alan~F Smeaton, Tomas~E Ward, and Graham Healy.
\newblock Neuroscore: A brain-inspired evaluation metric for generative
  adversarial networks.
\newblock {\em arXiv preprint arXiv:1905.04243}, 2019.

\bibitem{wang2019generative}
Zhengwei Wang, Qi She, and Tomas~E Ward.
\newblock Generative adversarial networks: A survey and taxonomy.
\newblock {\em arXiv preprint arXiv:1906.01529}, 2019.

\bibitem{yang2019gesture}
LI Yang, Jin HUANG, TIAN Feng, WANG Hong-An, and DAI Guo-Zhong.
\newblock Gesture interaction in virtual reality.
\newblock {\em Virtual Reality \& Intelligent Hardware}, 1(1):84--112, 2019.

\bibitem{zhang2018egogesture}
Yifan Zhang, Congqi Cao, Jian Cheng, and Hanqing Lu.
\newblock {EgoGesture}: {A} new dataset and benchmark for egocentric hand
  gesture recognition.
\newblock {\em IEEE Transactions on Multimedia}, 20(5):1038--1050, 2018.

\end{thebibliography}
}

\end{document}